\RequirePackage{fix-cm}
\RequirePackage[colorlinks,citecolor=blue,urlcolor=blue]{hyperref}

\documentclass[natbib, smallextended]{svjour3}       

\usepackage{minibox}

\usepackage{graphicx}
\usepackage{amsmath,amssymb,amsfonts}
\usepackage{empheq}
\usepackage[algoruled, lined, linesnumbered]{algorithm2e}
\usepackage{algcompatible}
\usepackage[utf8]{inputenc}
\usepackage{color}
\usepackage{lipsum}
\usepackage{comment}
\usepackage{placeins}
\usepackage{breqn}
\usepackage{pdflscape}
\usepackage{subfig}
\usepackage{pifont}
\usepackage{enumitem}
\usepackage[hang,flushmargin]{footmisc}
\usepackage[table]{xcolor}
\usepackage[flushleft]{threeparttable}
\newsavebox\CBox
\def\textBF#1{\sbox\CBox{#1}\resizebox{\wd\CBox}{\ht\CBox}{\textbf{#1}}}

\newenvironment{MyColorPar}[1]{\leavevmode\color{#1}\ignorespaces}{}

\newcommand{\bpi}{\boldsymbol{\pi}}
\newcommand{\bz}{\boldsymbol{z}}

\definecolor{orange}{rgb}{1,0.5,0}
\definecolor{redblue}{rgb}{1.0, 0.0, 1.0}

\usepackage{etoolbox}
\newcommand*{\affaddr}[1]{#1}
\newcommand*{\affmark}[1][*]{\textsuperscript{#1}}

\usepackage{booktabs}
\usepackage{multicol}
\usepackage{multirow,bigdelim}

\begin{document}

\title{A Non-Dominated Sorting Based Customized Random-Key Genetic Algorithm for the Bi-Objective Traveling Thief Problem}

\titlerunning{A NDS-BRKGA for the BI-TTP}        

\author{Jonatas B. C. Chagas \protect\affmark[1,2,*]\href{https://orcid.org/0000-0001-7965-8419}{\includegraphics[width=10pt,height=10pt]{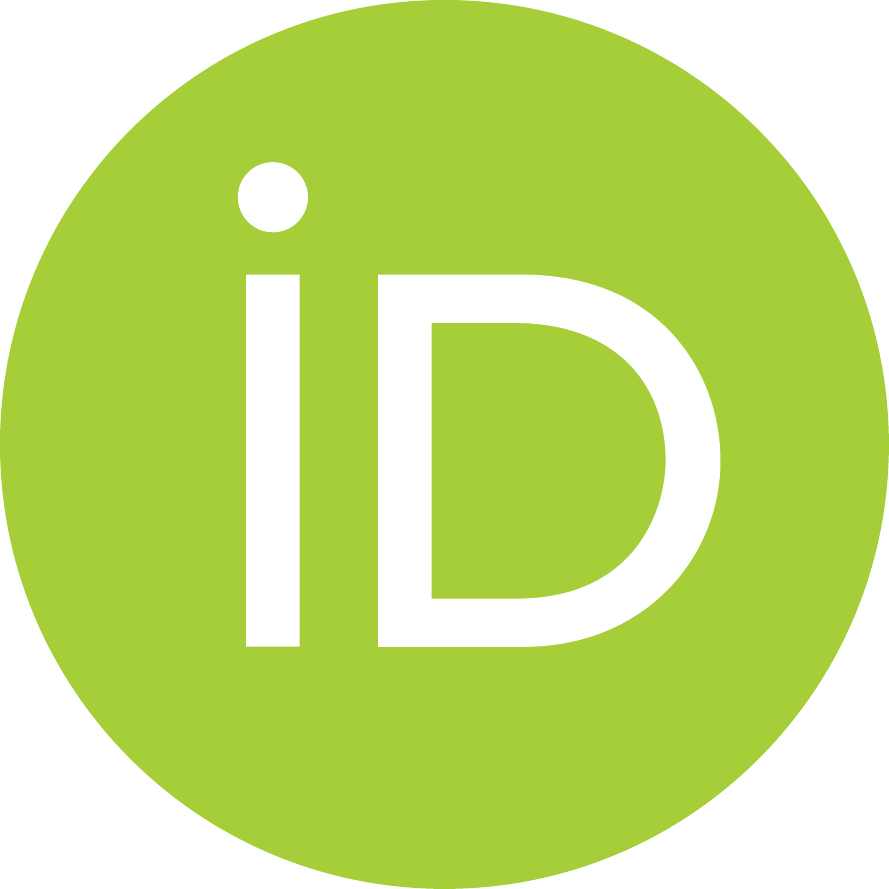}} 
\and \\
Julian Blank \affmark[3,*] \href{https://orcid.org/0000-0002-2227-6476}{\includegraphics[width=10pt,height=10pt]{orcid.pdf}} 
\and \\
Markus Wagner \affmark[4]\href{https://orcid.org/0000-0002-3124-0061}{\includegraphics[width=10pt,height=10pt]{orcid.pdf}}
\and \\
Marcone J. F. Souza \affmark[1]\href{https://orcid.org/0000-0002-7141-357X}{\includegraphics[width=10pt,height=10pt]{orcid.pdf}}   
\and \\
Kalyanmoy Deb \affmark[3] \href{https://orcid.org/0000-0001-7402-9939}{\includegraphics[width=10pt,height=10pt]{orcid.pdf}}
}

\authorrunning{Jonatas B. C. Chagas et al.} 

\institute{Jonatas B. C. Chagas \at
           \email{jonatas.chagas@iceb.ufop.br} 
           \and
           Julian Blank \at
           \email{blankjul@egr.msu.edu} 
           \and
           Markus Wagner \at
           \email{markus.wagner@adelaide.edu.au}
           \and
           Marcone J. F. Souza  \at
           \email{marcone@iceb.ufop.br}
           \and
           Kalyanmoy Deb \at
           \email{kdeb@egr.msu.edu}
           \\ \\
           \affaddr{\affmark[1] \mbox{Departamento de Computação, Universidade Federal de Ouro Preto, Ouro Preto, Brazil}}\\
            \affaddr{\affmark[2] Departamento de Informática, Universidade Federal de Viçosa, Viçosa, Brazil\\
            \affaddr{\affmark[3] Michigan State University, East Lansing, MI, USA}\\
            \affaddr{\affmark[4] School of Computer Science, The University of Adelaide, Adelaide, Australia}\\
            \affaddr{\affmark[*] These authors have contributed equally to this work}}
}

\date{Received: date / Accepted: date}

\maketitle

\begin{abstract}
In this paper, we propose a method to solve a bi-objective variant of the well-studied Traveling Thief Problem (TTP). The TTP is a multi-component problem that combines two classic combinatorial problems: Traveling Salesman Problem (TSP) and Knapsack Problem (KP). We address the BI-TTP, a bi-objective version of the TTP, where the goal is to minimize the overall traveling time and to maximize the profit of the collected items. Our proposed method is based on a biased-random key genetic algorithm with customizations addressing problem-specific characteristics. We incorporate domain knowledge through a combination of near-optimal solutions of each subproblem in the initial population and use a custom repair operator to avoid the evaluation of infeasible solutions. The bi-objective aspect of the problem is addressed through an elite population extracted based on the non-dominated rank and crowding distance. Furthermore, we provide a comprehensive study showing the influence of each parameter on the performance. Finally, we discuss the results of the BI-TTP competitions at \textit{EMO-2019} and \textit{GECCO-2019} conferences where our method has won first and second places, respectively, thus proving its ability to find high-quality solutions consistently.

\end{abstract}

%
%

\section{Introduction}

In optimization research, problems with different characteristics are investigated. To find an appropriate algorithm for a practical problem, often assumptions about characteristics are made, and then a suitable algorithm is chosen or designed. 
For instance, an optimization problem can have several components interacting with each other. Because of their interaction, they build an interwoven system~\citep{klamroth2017multiobjective} where interdependencies in the design and the objective space exist. An optimal solution for each component independently will, in general, not be a good solution for the interwoven optimization problem. Similarly, in multidisciplinary design optimization, various disciplines are linked with each other and influence the objective value(s). The optimization of an aircraft wing, for example, combines stress analysis, structural vibration, aerodynamics, and controls \citep{mdo}. Due to the interwovenness modifying a single decision variable is likely to affect all objective values. 

Such complex optimization problems usually require domain knowledge and a sufficient amount of computational resources to be invested. For this reason, many researchers prefer solving academic test problems to show the performance of their algorithms.
In order to provide an academic interwoven optimization test problem, the Traveling Thief Problem (TTP)~\citep{bonyadi2013travelling} was proposed in 2013, where two well-known subproblems, the Traveling Salesman Problem (TSP) and the Knapsack Problem (KP), interact with each other. As in the TSP problem, a so-called thief has to visit each city exactly once. In addition to just traveling, the thief can make a profit during its tour by stealing items and putting them in the rented knapsack. However, the thief's traveling speed decreases depending on the current knapsack weight, which then increases the rent that the thief has to pay for the knapsack.
Even though researchers have been investigating both subproblems for many years and a myriad of optimization algorithms have been proposed, the interaction of both problems with each other turned out to be challenging. 
The TTP problem seeks to optimize the overall traveling time and the profit made through stealing items. Most of the research focused on the single-objective problem, where the objectives are composed by using a weighted sum. To be more precise, the profit is reduced by the costs due to renting the knapsack. The costs are calculated by multiplying the overall traveling time by a renting rate.
However, because the traveling time and profit represent solutions with different trade-offs, the problem is bi-objective in its natural formulation.

In this article, we propose a Non-Dominated Sorting Biased Random-Key Genetic Algorithm (NDS-BRKGA) to obtain a non-dominated set of solutions for the BI-TTP. This problem is a bi-objective version of the TTP, where the goals are to minimize the overall traveling time and to maximize the profit of the collected items. The algorithm is based on the well-known evolutionary multi-objective optimization solution strategy NSGA-II~\citep{deb2002fast}, and the biased-random key encoding is used to deal with the mixed-variable nature of the problem.
The customization makes use of domain knowledge, which is incorporated by evolutionary operators. Our method uses existing solvers of the subproblems for the initial population. It maps a genotype to phenotype to deal with independent variables. Also, solutions detected as infeasible are repaired before they are evaluated. Moreover, we use a customized survival selection to ensure diversity preservation.

The remainder of this paper is structured as follows. In Section~\ref{sec:related}, we provide a brief review of the literature about the TTP. Afterward, we present a detailed description of the BI-TTP, as well as a solution example to demonstrate the interwovenness characteristic of the problem in Section~\ref{sec:problem}. In Section~\ref{sec:nsbrkga}, we describe our methodology to address the problem and present results evaluated on different test problems in Section~\ref{sec:computational-experiments}. Finally, the conclusions of the study are presented in Section~\ref{sec:conclusion}.

\section{Related Work}
\label{sec:related}

Thus far, many approaches have been proposed for the TTP. Most research so far considered the single-objective TTP formulation (TTP1) from \cite{bonyadi2013travelling}, which is typically the TTP variant that is referred to as \emph{the TTP}. 
The other variant (TTP2) considers two objectives and additionally, a value drop of items over time.
The multi-objective considerations of problems with interconnected components are becoming increasingly popular. 

For the single-objective TTP, a wide range of approaches has been considered, ranging from general-purpose iterative search-heuristics~\citep{polyakovskiy2014comprehensive}, co-evolutionary approaches~\citep{bonyadi2014socially,namazi2019cooperative}, memetic approaches~\citep{mei2014interdependence}, and swarm-intelligence based approaches~\citep{wagner2016stealing,Zouari2019antstpp}, to approaches with problem specific search operators~\citep{faulkner2015approximate}. On a higher, i.e., algorithmic level, estimation of distribution approaches have been customized~\citep{Martins2017ttpeda} and  hyper-heuristics~\citep{elyafrani2018hyperttp} have been explored.
\cite{wagner2018casestudy} provide a comparison of 21 algorithms for the purpose of portfolio creation. To better understand the effect of operators on the capability to find good solutions on a more fundamental level, \cite{elyafrani2018ttplandscape} and \cite{Wuijts2019ttpinvest} present in fitness-landscape analyses correlations and characteristics that are potentially exploitable.

Optimal approaches are rare but exist. \cite{neumann2017ttpPTAS} showed that the TTP with fixed tours can be solved in pseudo-polynomial time via dynamic programming taking into account the fact that the weights are integer. \cite{wu2017ttpexact} extended this to optimal approaches for the entire TTP, although their approaches are only practical for very small instances.

In general, the number of works on (static) multi-objective TTP formulations is significantly smaller. \citet{yafrani2017ttpemo} created an approach that generates diverse sets of solutions, while being competitive with the state-of-the-art single-objective algorithms; the objectives have been travel time and total profit of items.
\citet{wu2018evolutionary} considered a bi-objective version of the TTP; the objectives have been the weight and the TTP objective score. This hybrid approach makes use of the dynamic programming approach for fixed tours and then searches over the space of tours only. 

Moreover, researchers have participated in the BI-TTP competitions at the \textit{EMO-2019}\footnote{\url{https://www.egr.msu.edu/coinlab/blankjul/emo19-thief/}} and \textit{GECCO-2019}\footnote{\url{https://www.egr.msu.edu/coinlab/blankjul/gecco19-thief/}} conference. For these competitions, an intermediate version of TTP1 and TTP2 has been proposed. The problem description can be seen as a TTP1 with two objectives or TTP2 without a value drop of items over time. The same TTP variant was investigated by \citet{blank2017solvingBittp}.
The proposed problem definition aimed to build the bridge from TTP1 to TTP2 by having two objectives but not adding another complexity to the problem.

Recently, two dynamic formulations of the TTP have been devised by two groups and independent of each other: a single-objective variant~\cite{sachdeva2020dynamic} and a multi-objective variant~\cite{herring2020dynamic}. 
In the former, the authors considered the dynamic changes to the availability of cities and items. They have found that -- depending on the size of the instance, the magnitude of the change, and the algorithms in the portfolio -- it is preferable to either restart the optimization from scratch or to continue with the previously valid solutions. In the multi-objective article, the respective authors considered dynamic city locations, dynamic item availability, and dynamic item values. Experimentally, they investigated optimal solutions to TTP components and their recombination to generate diverse, composite populations for better responses to dynamic changes.

Furthermore, a more general discussion of a multi-objective approach to interconnected problems can be found in~\cite{klamroth2016interwoven}, and a more general discussion on the opportunities of multi-component problems can be found in~\cite{Bonyadi2019}.

\section{Bi-objective Traveling Thief Problem}
\label{sec:problem}

The TTP is a combinatorial optimization problem that consists of two interwoven problems, the TSP and KP. In the following, first, the two components are described independently, and then the interaction of the two subcomponents is shown.

In the TSP \citep{tsp_study} a salesman has to visit $n$ cities. The distances are given by a map represented as a distance matrix $A = (d_{ij})$ with $i,j \in \{0,..,n\}$. The salesman has to visit each city once and the result is a permutation vector $\bpi = (\pi_1, \pi_2, ..., \pi_n)$, where $\pi_i$ is the $i$-th city of the salesman.
The distance between two cities divided by a constant velocity $v$ (usually $v=1$) results in the traveling time for the salesman denoted by $f(\bpi)$.
The goal is to minimize the total traveling time of the tour: 

\begin{eqnarray}
\label{eqn:tsp}
\min \;\; f(\bpi) \;=\; \sum_{i=1}^{n-1} \frac{ d_{\pi_i, \pi_{i+1}}}{v} \; +  \;
\frac{ d_{\pi_n, \pi_{1}}}{v}  \\ [2mm]\notag
\text{s.t.} \qquad \bpi = (\pi_1, \pi_2, ..., \pi_n) \in P_n \\ \nonumber
\end{eqnarray}

There are $\frac{(n-1)!}{2}$ different tours to consider, 
if we assume that the salesman has to start from the first city 
and travels on a symmetric map, i.e., $d_{i,j} = d_{j,i},\; \forall (i, j) \in A$.

For KP~\citep{knp_survey} a knapsack has to be filled with items without violating the maximum weight constraint. Each item $j$ has a value $b_j \geq 0$  and a weight $w_j \geq 0$ where $j \in \{1, .., m\}$. The binary decision vector $\bz = (z_1, .., z_m)$ defines, if an item is picked or not. The aim is to maximize the profit $g(\bz)$:

\begin{eqnarray}
\max \;\; g(\bz) \;=\; \sum_{j=1}^{m}  z_j \, b_j \\[2mm] \notag 
\end{eqnarray}
\vspace{-1.1cm}
\begin{eqnarray}
\text{s.t.} \qquad & &\sum_{j=1}^m z_j \, w_j \leq Q \nonumber \\[2mm] 
& &\bz = (z_1, .., z_m) \in \mathbb{B}^m \nonumber \notag 
\end{eqnarray}

The search space of this problem is exponential concerning $n$ and contains $2^n$ possible combinations. However, the optimal solution can be obtained by using dynamic programming with a running time of $\mathcal{O}(mQ)$, which makes the complexity of the problem pseudo-polynomial, when the weights of items are integer values.

The traveling thief problem combines the above-defined subproblems and lets them interact with each other. The traveling thief can collect items from each city he/she is visiting. The items are stored in a rented knapsack carried by the thief. In more detail,
each city $\pi_i$ provides one or multiple items, which could be picked by the thief. There is an interaction between the subproblems:
The velocity of the traveling thief depends on the current knapsack weight $W$. It is calculated by considering all cities, which have been visited so far, and summing up the weights of all picked items. The weight at city $i$ given $\bpi$ and $\bz$ is calculated by:

\begin{equation}
    W(i, \bpi, \bz) = \sum_{k=1}^{i} \sum_{j=1}^{m} a_j(\pi_k)\; w_j z_j
\end{equation}

The function $a_j(\pi_k)$ is defined for each item $j$ and returns $1$ if the item could be stolen at city $\pi_k$ and $0$ otherwise.
The current weight of the knapsack influences the velocity.
When the thief picks an item, the weight of the knapsack increases, and therefore the velocity of the thief decreases.

The velocity $v$ is always in a specific range $v = (v_{\rm min}, v_{\rm max})$ and cannot be negative for a feasible solution.
Whenever the knapsack is heavier than the maximum weight $Q$, the capacity constraint is violated.

\begin{align}
 v(W) &=
  \begin{cases}
   \; v_{\rm max} - \frac{W}{Q} \cdot (v_{\rm max} - v_{\rm min}) & \text{if } W \leq Q \\
   \; v_{\rm min}        & \text{otherwise}
  \end{cases}
\end{align}

If the knapsack is empty, then the velocity is equal to $v_{\rm max}$.
Contrarily, if the current knapsack weight is equal to $Q$, the velocity is $v_{\rm min}$.

The traveling time of the thief is calculated by:

\begin{equation}
    f(\bpi, \bz) = \sum_{i=1}^{n-1} \frac{ d_{\pi_i, \pi_{i+1}}}{ v( W(i,\bpi,\bz) )  }
    \; + \; \; \frac{ d_{\pi_n, \pi_{1}}}{ v( W(n,\bpi,\bz) )}  \\
\end{equation}

The calculation is based on TSP, but the velocity is defined by a function instead of a constant value.
This function takes the current weight, which depends on the index $i$ of the tour.
The current weight, and therefore also the velocity, change on tour by considering the picked items defined by $\bz$.
In order to calculate the total tour time, the velocity at each city needs to be known.
For calculating the velocity at each city, the current weight of the knapsack must be given.
Since both calculations are based on $\bz$, i.e., the knapsack subproblem solution, it is challenging to solve the problem to optimality. 
Such problems are called interwoven systems as the solution of one subproblem highly depends on the solution of the other subproblems.

After this preliminary presentation, we finally formalize the TTP as follows.

\begin{eqnarray}
    \min & & f(\bpi, \bz) \;=\; \sum_{i=1}^{n-1} \frac{ d_{\pi_i, \pi_{i+1}}}{ v( W(i,\bpi,\bz) ) }  \; + \; \; \frac{ d_{\pi_n, \pi_{1}}}{ v( W(n,\bpi,\bz) )}\\[2mm] 
    \max & & g(\bz) \;=\; \sum_{j=1}^{m}  z_j \, b_j \nonumber
\end{eqnarray}
\begin{eqnarray}
    \text{s.t.} \qquad & &\bpi = (\pi_1, \pi_2, ..., \pi_n) \in P_n \nonumber\\[1mm]  
    & &\pi_1 = 1 \nonumber\\[1mm]  
    & &\bz = (z_1, .., z_m) \in \mathbb{B}^m \nonumber\\ 
    & &\sum_{j=1}^m z_j \, w_j \leq Q \nonumber\\  \nonumber
\end{eqnarray}

\begin{figure}[!ht]
\centering
\includegraphics[width=0.8\textwidth]{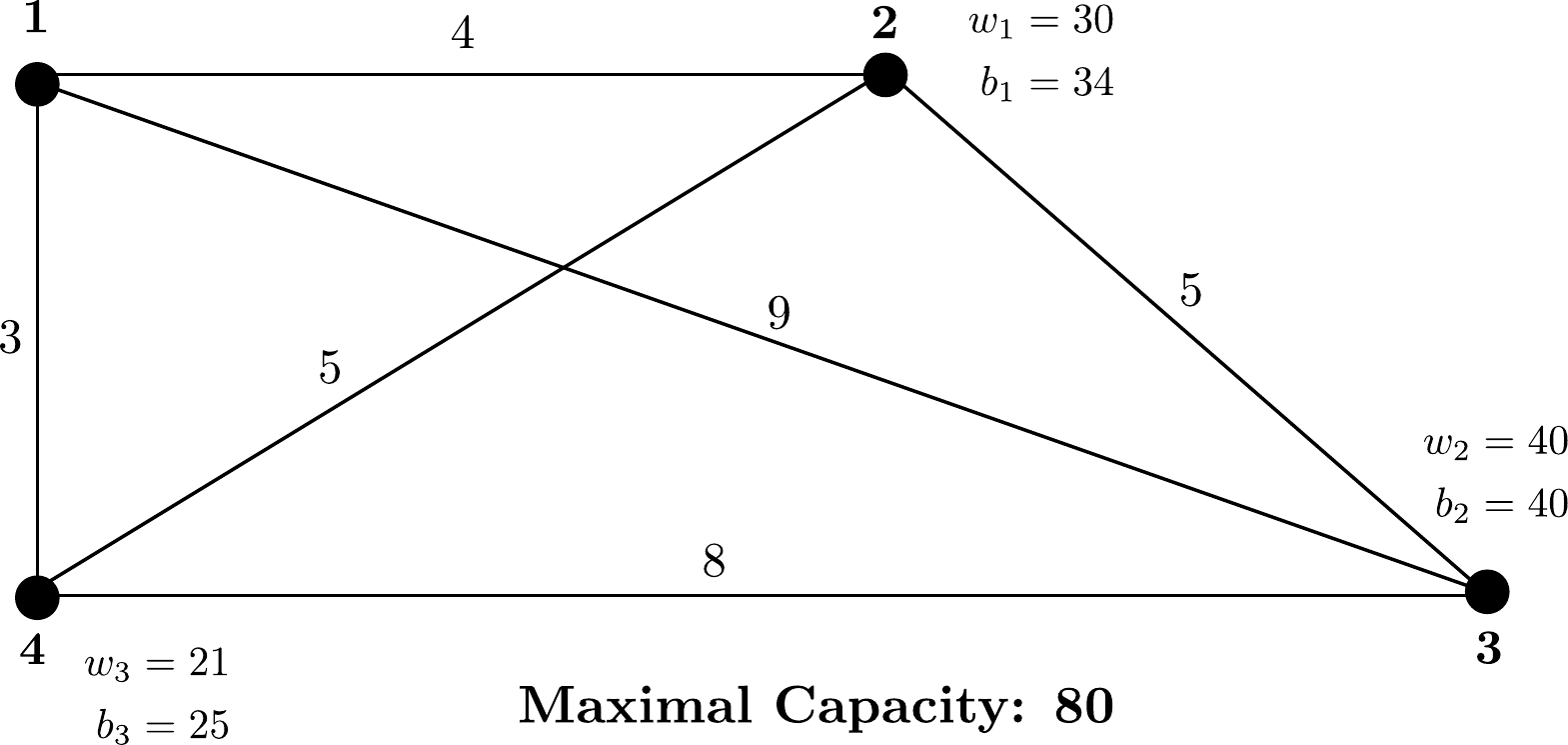}
\caption{Exemplary traveling thief problem instance.}
\label{fig:unique_hyperplane}
\end{figure}

In order to illustrate the equations and interdependence, we present an example scenario here (see Figure~\ref{fig:unique_hyperplane}). 
The thief starts from city 1 and has to visit city 2, 3, 4 exactly once and to return to city 1.
In this example, each city provides one item, and the thief must decide whether to steal it or not.

A permutation vector, which contains all cities exactly once, and a binary picking vector are needed to calculate the objectives.
Even though this is a very small example with four cities and three items, the total solution space consists of $(n-1)! \cdot 2^m = 6 \cdot 8 = 48$ combinations.

In order to understand how the objectives are calculated, an example hand calculation for the tour [1,3,2,4] and the packing plan [1,0,1] is done as follows. 
The thief starts with the maximum velocity, because the knapsack is empty. He begins its tour at city $1$ and picks no item there. 
For an empty knapsack $W(1,\bpi,\bz) = 0$ the velocity is $v(0) = v_{\rm max} = 1.0$. The distance from city $1$ to city $3$ is $9.0$ and the thief needs $9.0$ time units.
At city $3$ the thief will not pick an item and continue to travel to city $2$ with $W(2,\bpi,\bz) = 0$ and therefore with $v_{\rm max}$ in additional $5.0$ time units.
Here he picks item $1$ with $w_1 = 30$ and the current weight becomes $W(2,\bpi,\bz) = 30$,
which means the velocity will be reduced to $v(30) = 1.0 - (\frac{1.0 - 0.1}{80}) \cdot 30 = 0.6625$.
For traveling from city $2$ to city $4$ the thief needs the distance divided by the current velocity $\frac{5.0}{0.6625} \approx 7.5472$.
At city $4$ he picks item $3$ with $w_3 = 21$ and the current knapsack weight increases to $W(4,\bpi,\bz) = 30 + 21 = 51$.
For this reason the velocity decreases to $v(51) = 1.0 - (\frac{1.0 - 0.1}{80}) \cdot 51 = 0.42625$.
For returning to city $1$ the thief needs according to this current speed $\frac{3.0}{0.42625} \approx 7.0381$ time units.
Finally, we sum up the time for traveling from each city to the next $\sum_{k=1}^{4} t_{\pi_k, \pi_{k+1}} = 9 + 5 + 7.5472 + 7.0381 = 28.5853$ to calculate the whole traveling time.

\begin{table}[!htbp]
\centering
\caption{Hand calculations for [0,2,1,3] [0,1,0,1] on the example scenario.} 
\begin{tabular}{p{10pt}|ccccc|c}
i & $\pi_i$ & $W(i,\bpi,\bz)$ & $v(W(i,\bpi,\bz))$ & $d_{\pi_i, \pi_{i+1}}$ & $t_{\pi_i, \pi_{i+1}}$ & $\sum$ \\ \hline
1 & 1 & 0 & 1 & 9 & 9 & - \\
2 & 3 & 0 & 1 & 5 & 5 & 9 \\
3 & 2 & 30& 0.6625  & 5 & 7.5472 & 14 \\
4 & 4 & 51 & 0.42625  & 3& 7.0381 & 21.547 \\ \hline
5 & 1 & - & - & - & - & \textBF{28.585}  \\
\end{tabular}
\end{table}

The final profit is calculated by summing up the values of all items which is $34 + 25 = 59$. Consequently, the TTP solution [1,3,2,4] [1,0,1] is mapped to the point $(28.59, 59.0)$ in the bi-objective space.

Below all Pareto-optimal solutions of this example are listed. The Pareto front contains $8$ solutions, of which two has the same minimum $f(\pi,z)$ value. The solution for the hand calculation is highlighted in bold.

\begin{table}[!htbp]
\centering
\caption{Pareto front of the example scenario.}
\setlength{\tabcolsep}{20pt}
\begin{tabular}{cc|cc}
\toprule
$\bpi$ & $\bz$ & $f(\bpi,\bz)$ & $g(\bz)$ \\
\midrule
\text{[1, 2, 3, 4]} & [0, 0, 0] & 20.0 & 0.0\\
\text{[1, 4, 3, 2]} & [0, 0, 0] & 20.0 & 0.0\\
\text{[1, 2, 3, 4]} & [0, 0, 1] & 20.93 & 25.0\\
\text{[1, 4, 3, 2]} & [1, 0, 0] & 22.04 & 34.0\\
\text{[1, 4, 3, 2]} & [0, 1, 0] & 27.36 & 40.0\\
\textBF{[1, 3, 2, 4]} & \textBF{[1, 0, 1]} & \textBF{28.59} & \textBF{59.0}\\
\text{[1, 2, 3, 4]} & [0, 1, 1] & 33.11 & 65.0\\
\text{[1, 4, 3, 2]} & [1, 1, 0] & 38.91 & 74.0\\
\bottomrule
\end{tabular}
\label{tbl:example_front}
\end{table}

\begin{figure}[!ht]
\centering
\includegraphics[width=0.95\linewidth]{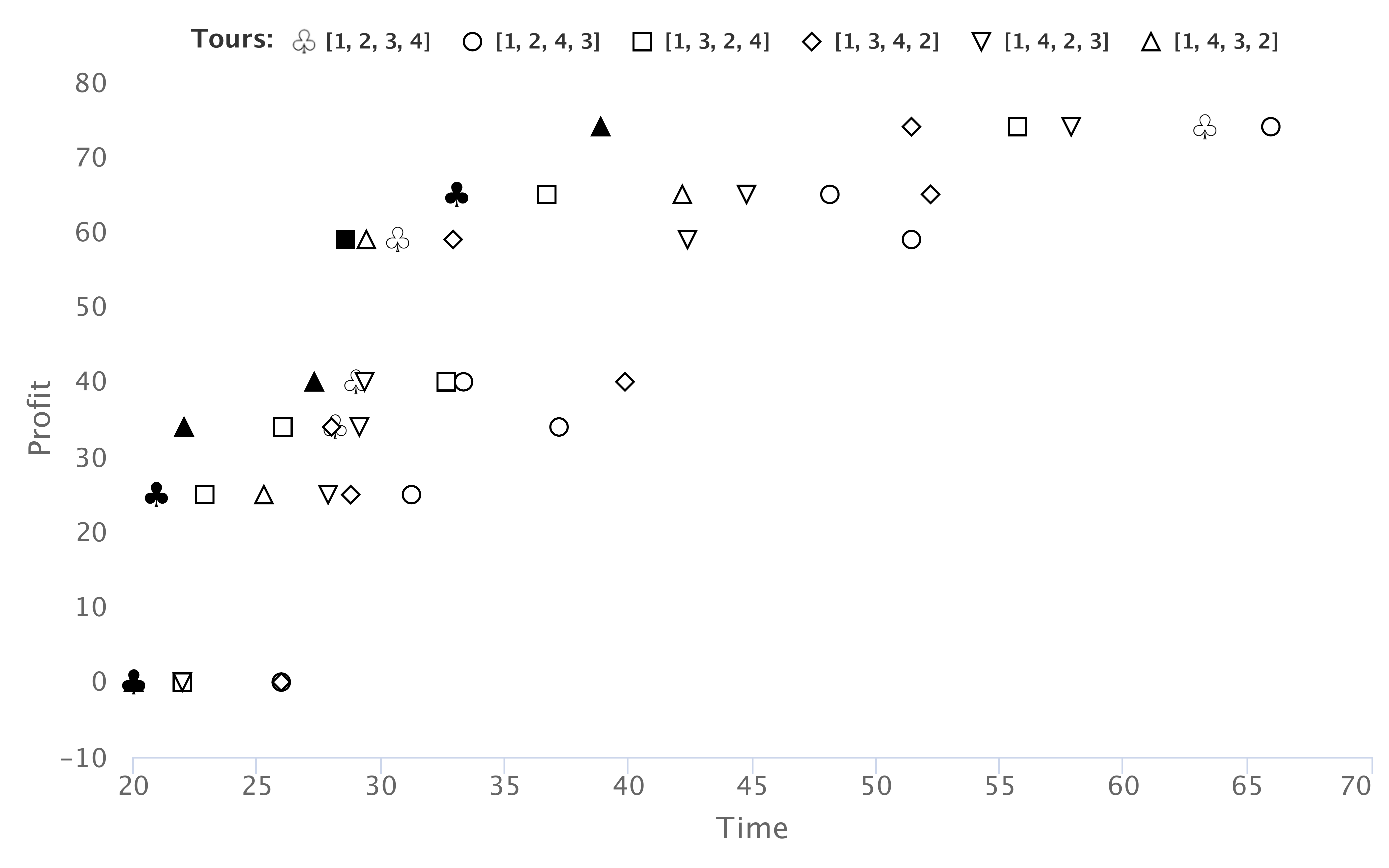}
\caption{Exemplary traveling thief problem instance. Bold symbols indicate non-dominated solutions.}
\label{fig:example_front}
\end{figure}

Figure~\ref{fig:example_front} shows the objective space by highlighting different tours with different markers. The non-dominated solutions are emphasized by black markers. We can observe that different Pareto-optimal solutions can have different underlying tours. In addition, we can see that for each solution $s$ where no item is picked, there is another solution $s'$ with its tour symmetric to the tour of $s$, and, consequently, both solutions $s$ e $s'$ have the same traveling time, once we consider that the thief travels on a symmetric map. Also, no solution with a tour $\nabla=[1, 4, 2, 3]$ exists in the final non-dominated set.

\section{A Customized Non-dominated Sorting Based Genetic Algorithm with Biased Random-Key Encoding}
\label{sec:nsbrkga}

Genetic algorithms (GAs) provide a good starting point because almost no assumptions about the problem properties are made. GAs are highly customizable, and the performance can be improved through defining/redefining the evolutionary operators. For the BI-TTP, we propose a Non-Dominated Sorting Biased Random-Key Genetic Algorithm (NDS-BRKGA), which combines two classical evolutionary metaheuristics: Biased Random-Key Genetic Algorithm (BRKGA)~\citep{gonccalves2011biased} and Non-Dominated Sorting Genetic Algorithm II (NSGA-II)~\citep{deb2002fast}. Both concepts come together to address the following characteristics of the BI-TPP: 

\begin{enumerate}[label=(\roman*), leftmargin=13mm]
\setlength\itemsep{2mm}

\item \textbf{Existing solvers for each subproblem}: Both subproblems, TSP and KP, have been studied for decades, and good solvers for each problem exist. We incorporate this domain knowledge by using a heuristic-based initial population by combining near-optimal solutions of each subproblem. In our initial population, we seek to preserve a high diversity among individuals, in order not to lead our algorithm to premature convergence.

\item \textbf{Maximum capacity constraint:} Through a repair operation before any evaluation of an individual, the domain knowledge can be incorporated to avoid the evaluation of infeasible solutions. An effective repair allows the algorithm to search only in the feasible space. 

\item \textbf{Heterogeneous variable types:} A tour (permutation) and a packing plan (binary decision vector) need to be provided to evaluate a solution. Both variables are linked with each other. Handling different types of variables can be challenging; therefore, we introduce a real-valued genotype by using the biased-random key principle. This allows applying traditional evolutionary recombination operators on continuous variables.

\item \textbf{Bi-objective:} The traveling time of the thief is supposed to be minimized, and the profit to be maximized. We consider both conflicting objectives at a time by using the non-dominated sorting and crowding distance in the survival selection. This ensures the final population contains a set of non-dominated solutions with a good diversity in the objective space.

\end{enumerate}

In the remainder of this section, we first explain the overall procedure and then the role of each criterion mentioned above. 

\vspace{3mm}
\noindent \textbf{Overview}
\vspace{1mm}

\begin{figure}[!ht]
\centering
\includegraphics[width=0.9\linewidth]{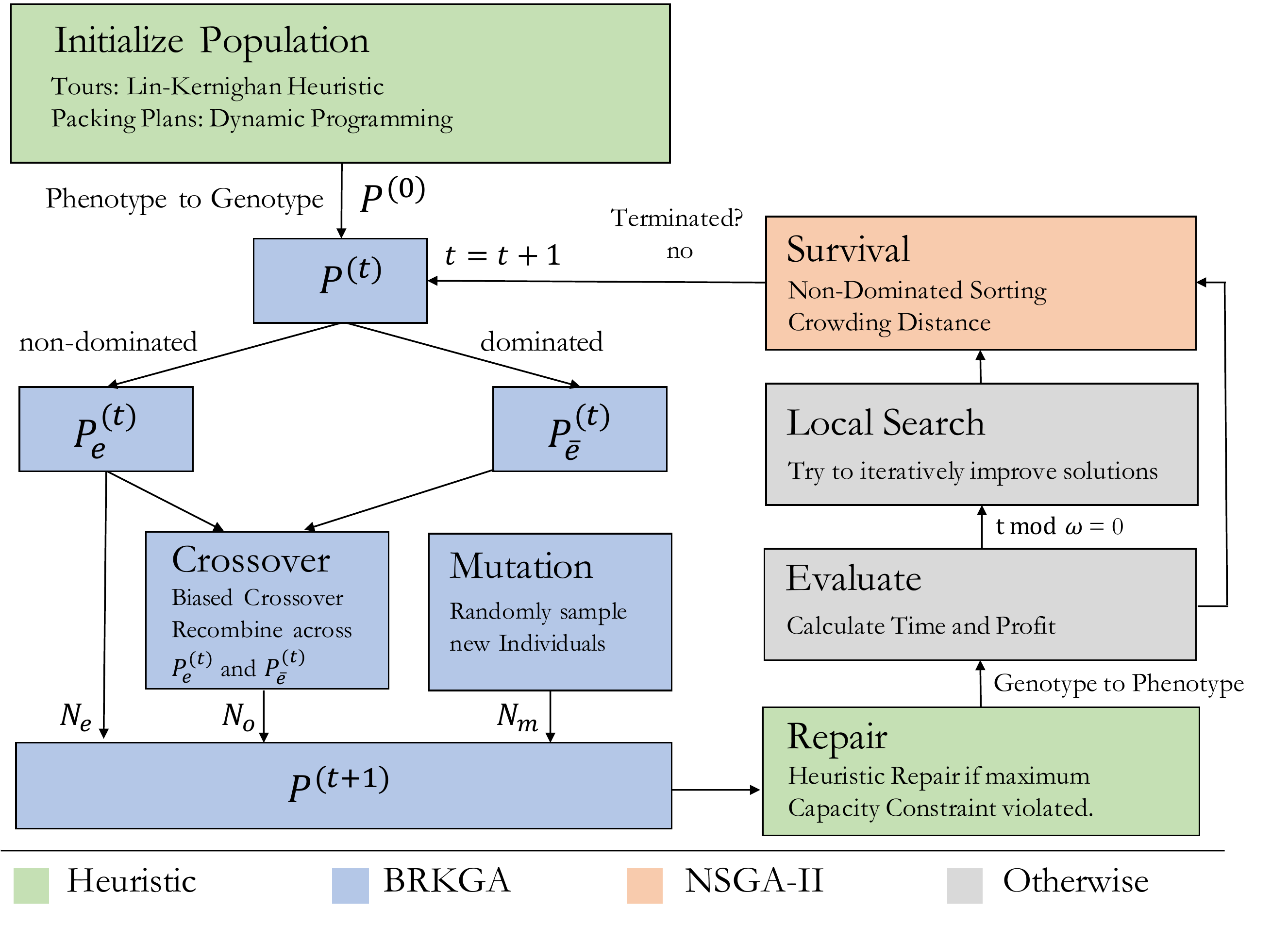}
\caption{NDS-BRKGA: A customized genetic algorithm.}
\label{fig:algorithm}
\end{figure}

\noindent Figure~\ref{fig:algorithm} illustrates the overall procedure of NDS-BRKGA. At first, we generate the initial population using efficient solvers for the subproblems independently. Afterward, we combine the optimal or near-optimal solutions for both subproblems and convert them to their genotype representation, which results in the initial population. For the purpose of mating, the population is split into an elite population $P_e^{(t)}$ and non-elite population $P_{\bar{e}}^{(t)}$. The individuals for the next generations $P^{(t+1)}$ are a union of the elite population $P_e^{(t)}$ directly, the offspring of a biased crossover and mutant individuals. In case an individual violates the maximum capacity constraint, we execute a repair operation. Then, we convert each individual to its corresponding phenotype and evaluate it on the problem instance. In order to insert an explicit exploitation phase in our algorithm, we apply at some evolutionary cycles a local search procedure in some elite individuals. Finally, the survival selection is applied, and if the termination criterion is not met, we increase the generation counter $t$ by one and continue with the next generation. In the following, we describe the purpose of each of the design decisions we have made and explain what role it plays during a run of the algorithm.

\vspace{3mm}
\noindent \textbf{Genotype to phenotype decoding}
\vspace{1mm}

\noindent In order to facilitate the exploration of the BI-TTP solution space, we represent the genotype of each individual as a vector of random-keys, which is a vector of real numbers between the interval [0,1]. It is an indirect representation that allows us to navigate in the feasible solution space of any optimization problem through simple genetic operators.  This representation strategy has been successfully applied to several complex optimization problems \citep{gonccalves2012parallel, resende2012biased, gonccalves2013biased, lalla2014biased, gonccalves2015biased, santos2018thief}.

\begin{figure}[!ht]
    \centering
    \includegraphics[width=0.8\linewidth]{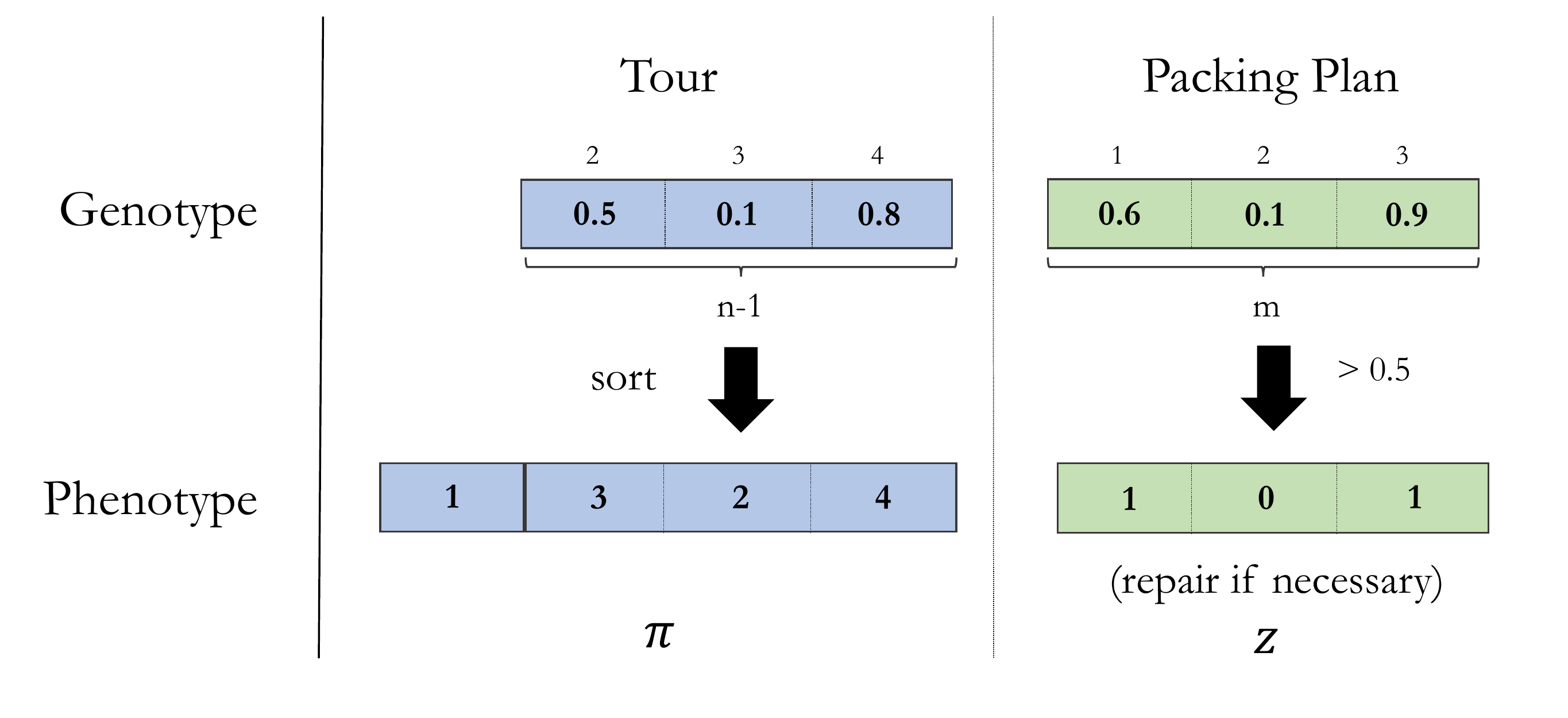}
    \caption{Chromosome structure: Genotype to phenotype mapping.}
    \label{fig:genotype}
\end{figure}

Because this representation is independent of the problem addressed, a deterministic procedure is necessary to decode each individual to a feasible solution of the problem at hand, i.e., an algorithm that decodes a genotype to its respective phenotype. In Figure~\ref{fig:genotype}, we illustrate the genotype and phenotype structure for the BI-TTP. The structure can be divided into two parts, the tour, and the packing plan. The tour needs to be decoded to a permutation vector. It is known that the thief is starting from city $1$ and, therefore, the order of the remaining $n-1$ cities needs to be determined. To achieve this, the sorting of the random key vector with length $n-1$ forms a permutation from $1$ to $n-1$. Then, each value is increased by $1$ to shift the permutation from $2$ to $n$. By appending this permutation to the first city, the tour is decoded to its phenotype. The packing plan needs to be decoded to a binary decision vector of length $m$. The decision of whether to pick an item or not is made based on the value of the biased random key, which has the same length. If the corresponding value is larger than $0.5$, the item is picked up, otherwise not. A exemplary decoding of the biased random key vector $[0.5, 0.1, 0.8, 0.6, 0.1, 0.9]$ (see Figure~\ref{fig:genotype}) would be the following: First separate the genotype into two parts $[0.5, 0.1, 0.8]$ and $[0.6, 0.1, 0.9]$. Then, sort the first vector and increase each value by $1$ results in the permutation $[3,2,4]$. By appending it this vector to the first city, the tour is $\bpi=[1,3,2,4]$. For the second part, for each value in $[0.6, 0.1, 0.9]$ we set the bit if it is larger than $0.5$ which results in $[1,0,1]$. Note that this example decodes to the variable used for our hand-calculation in Section~\ref{sec:problem}. Moreover, the decoding is a many-to-one mapping, which means different genotypes can represent the same phenotype.

\vspace{3mm}
\noindent \textbf{Repair operator}
\vspace{1mm}

\noindent According to the decoding procedure previously described, a genotype can generate an infeasible phenotype concerning the packing plan. It occurs when the total weight of the picked items is higher than the maximum limit of the knapsack. In order to repair an infeasible genotype, we apply an operator that removes items from the packing plan until it becomes feasible. In this repair operator, we give preference to keeping items collected last; since this way, the thief can travel faster at the beginning of its journey. Therefore, we first remove all items collected from the first city visited by the thief. If the removal of these items makes the packing plan feasible, the repair operator is finished; otherwise, we repeat the previous step considering all items of the next city visited by the thief. This process repeats until the weight of all remaining items in the packing plan does not exceed the limit of the knapsack. We also repair the genotype to avoid propagating non-feasibility throughout the evolutionary process. For this purpose, we should assign any real number less than 0.5 to every random-key that references an item that has not been collected. In our implementation, we have used the number zero.

\vspace{3mm}
\noindent \textbf{Initial population}
\vspace{1mm}

\noindent We use a biased initial population to incorporate domain knowledge into the genetic algorithm. Because both subproblems of the BI-TTP are well-studied, we make use of existing algorithms to generate a good initial population. We maintain a population $\mathcal{P}$ of $N$ individuals throughout the evolutionary process. To create the initial population $\mathcal{P}^{(0)}$, we combine the tour found by TSP and the packing plan by KP solvers. To be not too biased to near-optimal solutions of each subproblem, those combinations represent only a small fraction of the entire population. Because the corresponding solvers provide the phenotype presentation, we convert them to their genotype representation to be able to apply evolutionary operators later on. We complete the population $\mathcal{P}^{(0)}$ by adding randomly created individuals to it, where each random-key is generated independently at random in the real interval~$[0, 1]$.

\begin{algorithm}
\makeatletter
\newcommand{\algorithmfootnote}[2][\footnotesize]{%
  \let\old@algocf@finish\@algocf@finish
  \def\@algocf@finish{\old@algocf@finish
    \leavevmode\rlap{\begin{minipage}{\linewidth}
    #1#2
    \end{minipage}}%
  }%
}
\DontPrintSemicolon
\SetKwData{Left}{left}
\SetKwData{Up}{up}
\SetKwFunction{FindCompress}{FindCompress}
\SetKwInOut{Input}{input}
\SetKwInOut{Output}{output}
\BlankLine
$\bpi' \gets $ solve the TSP component \tcp*{LKH algorithm} \label{alg:initial_population_lkh}
$\bpi'' \gets \{\;\pi_{1}, \pi_{n}, \pi_{n-1}, ..., \pi_{2}\;\}$ \tcp*{symmetric $\bpi'$} \label{alg:initial_population_symmetric_route}
$\bz' \gets $ solve the KP component \tcp*{GH + DP algorithm} \label{alg:initial_population_ghdp}
\BlankLine
$\bz'' \gets \varnothing$ \label{alg:initial_population_create_individuals_begin} \\
$\mathcal{S} \gets \{\;\zeta(\bpi', \bz''),\, \;\zeta(\bpi'', \bz'')\}$ \label{alg:initial_population_first_two_individuals} \\
\Repeat{$z' = \varnothing$} { \label{alg:initial_population_begin_loop}
    $i \gets $ select item $i \in z'$ with the highest $p_{i}/w_{i}$ rate \\
    $\bz'' \gets \bz'' \;\cup\; \{\;i\;\}$ \\
    $\bz' \gets \bz' \;\setminus\; \{\;i\;\}$ \\
    \eIf{\upshape $\;(\bpi', \bz'') \text{ is not dominated by } (\bpi'', \bz'')$} {
        $\mathcal{S} \gets \mathcal{S} \;\cup\; \{\;\zeta(\bpi', \bz'')\}$
    }
    {
        $\mathcal{S} \gets \mathcal{S} \;\cup\; \{\;\zeta(\bpi'', \bz'')\}$
    }
} \label{alg:initial_population_create_individuals_end}
\BlankLine
$\mathcal{A} \gets $ select randomly a set of $\alpha \times N$ individuals from $\mathcal{S}$ using a uniform distribution \label{alg:initial_population_uniform_individuals} \\
$\mathcal{B} \gets $ generate a set of $(1-\alpha) \times N$ random individuals \label{alg:initial_population_random_individuals} \\
$\mathcal{P}^{(0)} \gets \mathcal{A} \;\cup\; \mathcal{B}$ \label{alg:initial_population_merge} \\
\Return $\mathcal{P}^{(0)}$ \label{alg:initial_population_return}
\caption{Generate initial population}
\label{alg:initial_population}
\algorithmfootnote{$\zeta(\bpi, \bz)$ encodes the BI-TTP solution $(\bpi, \bz)$ to a vector of random-keys.}
\end{algorithm}%

In Algorithm \ref{alg:initial_population}, the required steps to create the initial population $\mathcal{P}^{(0)}$ are described in more detail. At first, we use the Lin-Kernighan Heuristic (LKH) \citep{lin1973effective} for solving the TSP component (Line~\ref{alg:initial_population_lkh}). We consider the symmetrical tour found by LHK (Line~\ref{alg:initial_population_symmetric_route}). As we consider that the thief travels on a symmetric map, where both these tours result in the same overall traveling time. Note that achieving near-optimal TSP tours is not a guarantor for near-optimal TTP solutions, and it has been observed that slightly longer tours have the potential to yield overall better TTP solutions \citep{wagner2016stealing, wu2018evolutionary}. However, we observed that near-optimal TSP tours combined with KP packing with lighter items generate BI-TTP solutions very close to the Pareto front regarding the traveling time objective. 

Next, we apply a two-stage heuristic algorithm, which has been developed by us for solving the KP component (Line \ref{alg:initial_population_ghdp}). We named this two-stage heuristic algorithm GH+DP because it combines a Greedy Heuristic (GH) with classical Dynamic Programming (DP) for solving the knapsack problem. The GH+DP algorithm starts by sorting all $m$ items according to the profit/weight ratio in non-increasing order. It then proceeds to insert the first $m'$ items such that the total weight $\sum_{i = 1}^{m'} w_{i}$ is not greater than $Q - \delta$ where delta is a parameter of our method. Next, it uses the classic dynamic programming algorithm \citep{toth1980dynamic} for solving the smaller KP considering the last $m - m'$ items and a knapsack of capacity $Q - \sum_{i = 1}^{m'} w_{i}$. There is no guarantee that near-optimal KP packing plans generate near-optimal TTP solutions. However, in contrast to single-objective approaches, we have observed that we can generate BI-TTP solutions close to the optimal profit objective by combining near-optimal KP packing plans with efficient TSP tours (see Section~\ref{sec:comparisonSOTTP}).

Afterward, we combine TSP and KP solutions to create new individuals (Line \ref{alg:initial_population_create_individuals_begin} to \ref{alg:initial_population_create_individuals_end}). Note that we first create two individuals (Line \ref{alg:initial_population_first_two_individuals}) from the tour (and its symmetric tour) found by LKH and from the empty knapsack solution. Next, iteratively, we create new non-dominated individuals so that at each iteration a single individual is created from the TSP solutions previously considered and also from a partial solution of the KP solution found by GH+DP algorithm. After creating all individuals, we select only a subset of them to compose the initial population. We randomly select $\alpha \times N$ ($\alpha$ is a parameter with its value between 0 and 1) individuals uniformly distributed from all individuals generated (Line \ref{alg:initial_population_uniform_individuals}), then we generate $(1-\alpha) \times N$ random individuals (Line \ref{alg:initial_population_random_individuals}) in order to complete the initial population, which is returned at the end of algorithm (Line \ref{alg:initial_population_return}).
 
\vspace{3mm}
\noindent \textbf{Elite and non-elite population}
\vspace{1mm}

\begin{figure}[!ht]
    \centering
    \includegraphics[width=0.8\linewidth]{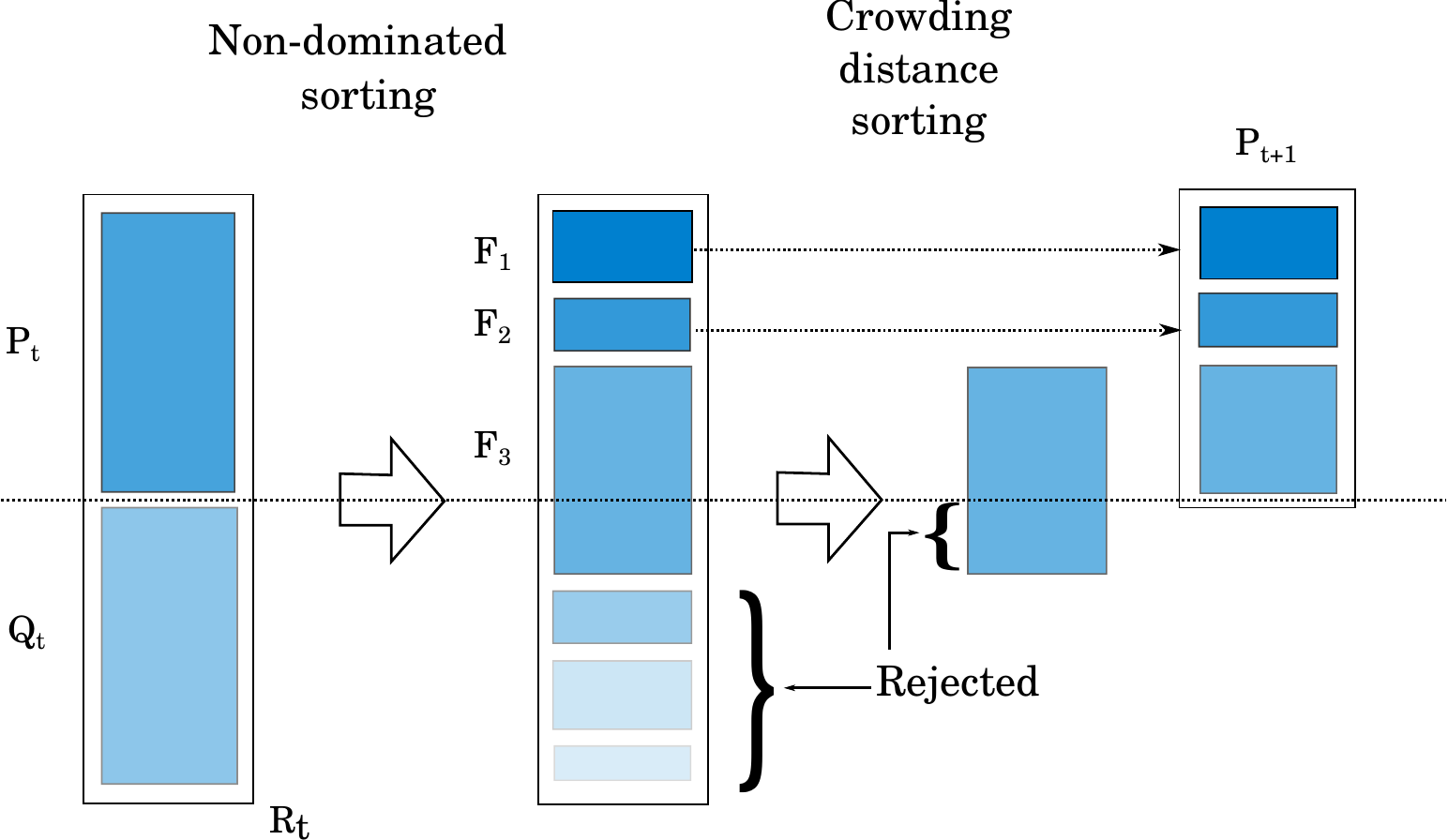}
    \caption{NSGA-II survival selection (Illustration based on~\cite{deb2002fast}).}
    \label{fig:nsga2}
\end{figure}

\noindent It is a common strategy of multi-objective optimization algorithms to give more importance to non-dominated solutions in the population during the recombination and environmental survival~\cite{Chand2015manyemo}. We split the population into two groups: the elites and non-elites. The number of elites is defined beforehand by the parameter $N_e$. 
We use the survival selection of NSGA-II~\citep{deb2002fast} as a splitting criterion (see  Figure~\ref{fig:nsga2}).
The current population $P^{(t)}$ and the offspring $Q^{(t)}$ are merged together and non-dominated sorting is applied. The outcome is a number of fronts $F_1, F_2, \ldots, F_L$, each of which is a set of individuals. Because the survival selection requires to select only $|P^{(t)}|$ individuals from the merged population, it might be the case that a front needs to be split into surviving and non-surviving individuals. In our example, $F_1$ and $F_2$ are surviving individuals because of their non-domination criterion. However, $F_3$ needs to be split. Therefore, a second criterion, crowding distance, is introduced. Based on the distance to neighboring individuals in the objective space, a crowding distance metric is calculated and assigned to each individual.

We use the non-dominated sorting and crowding distance to incorporate elitism. As is usually done, before calculating the crowding distance, we normalize the objectives in order to avoid a possible higher influence of a single objective. The number of elite individuals $N_e$ is determined by executing the NSGA-II survival on our current population $P^{(t)}$ with the goal to let $N_e$ individuals survive. The resulting survivors are added to the group of elites $P^{(t)}_e$ and the remaining to the group of non-elites $P^{(t)}_{\bar{e}}$.

\vspace{3mm}
\noindent \textbf{Biased crossover}
\vspace{1mm}

\noindent With the purpose of diversity preservation, we apply a biased crossover in order to create new offspring individuals. It is common practice when random-keys are used as a genotype representation.
The biased crossover operator involves two parents. The first is randomly selected from the elite population $P^{(t)}_e$ and the second randomly from the whole population $P^{(t)}$. Moreover, the biased crossover operator has a parameter $\rho_{e}$, which defines the probability of each random-keys of the first parent (it always belongs to the elite population) to be inherited by the offspring individual. More precisely, from an elite parent $a$ and another any parent $b$, we can generate an offspring $c$ according to the biased crossover as follows:

\begin{equation}
c_{i} \gets
\begin{cases}
\;a_{i} \quad \quad \text{if} \; \; \textsc{rand}\,(0,\,1)\;\leq\;\rho_{e}\\
\;b_{i} \quad \quad \text{else}
\end{cases}
\; \; \forall i \in \{1, 2, ..., n-1+m\} \nonumber
\end{equation}

\noindent where $a_{i}$, $b_{i}$ and $c_{i}$ are, respectively, the $i$-th random-key of individuals $a$, $b$ and $c$.

\vspace{3mm}
\noindent \textbf{Mutant individuals}
\vspace{1mm}

\noindent As in BRKGAs, we are not using any mutation operators, which are commonly used in most GAs \citep{mitchell1998introduction}. In order to maintain diversity in the population, we use so-called mutant individuals. Mutant individuals are simply randomly created individuals where each random-key is sampled from a uniform distribution in the $[0,1]$ range.

\vspace{3mm}
\noindent \textbf{Survival}
\vspace{1mm}

\noindent The population of the next generation $\mathcal{P}^{(t+1)}$ is formed based on the current population $\mathcal{P}^{(t)}$. The survival is the union of three different groups of individuals: 

\begin{enumerate}[label=(\roman*), leftmargin=13mm]
\setlength\itemsep{2mm}

\item \textbf{Elite population:} Part of the survival is based on elitism. Before the mating, a sub-population $P_e$ of $N_e$ individuals are selected as elites according to the NSGA-II criteria. This means non-dominated sorting determines the rank of each solution, and then for each non-dominated front, the crowding distance is calculated. In the case of a tie, two solutions are ordered randomly. Based on this absolute ordering, we pick $N_e$ individuals and directly copy them to $\mathcal{P}^{(t+1)}$.

\item \textbf{Mutant individuals:} In order to maintain a high diversity, a population $P_m$ of $N_{m}$ mutant individuals are added to $\mathcal{P}^{(t+1)}$. The strategy of keeping a separate set with mutants introduces a diversity of offspring during the mating. Otherwise, the recombination would be biased towards elites in the population, and a premature converge through a loss of diversity is likely.

\item \textbf{Offsprings from biased crossover}: To complete the number of individuals in the next population $\mathcal{P}^{(t+1)}$, $N - N_e - N_m$ individuals are generated and added on it through mating by the biased crossover operator. The biased crossover chooses one parent from the elites and another one from the non-elites. This mating is even more biased towards the elite population than the traditional binary tournament crossover used in NSGA-II, because it forces for each mating a non-dominated solution to participate in it.

\end{enumerate}

Finally, the surviving individuals are obtained by merging these three sets together $\mathcal{P}^{(t+1)} = P_e \cup P_m \cup P_o$. The survival is partly based on elitism through letting $P_e$ survive for sure, but also adds two more diverse groups through evolutionary operators. 

\vspace{3mm}
\noindent \textbf{Local search}
\vspace{1mm}

At some evolutionary cycles, we apply an exploitation procedure of the search space by modifying the genotypes of some individuals to enhance the fitness of the current population. This methodology is commonly applied to traditional genetic algorithms in order to balance the concepts of exploitation and exploration, which are aspects of effective search procedures~\citep{neri2012memetic}.
Genetic Algorithms (GAs) with exploitation procedure are known and widely referenced as Memetic Algorithms (MAs). According to \cite{krasnogor2005tutorial}, MAs have been demonstrated to be more effective than traditional GAs for some problem domains, especially for combinatorial optimization problems.

In NDS-BRKGA, the local search is only applied to some percentage of the population and consists of two phases. First, the tour $\bpi$ is considered separately and the permutation is modified; Second, it considers only the packing plan $\bz$ and through bit-flips items are either removed or added.
In Algorithm~\ref{alg:self_improvement} we describe the exploitation phase in more detail.
In order to ensure a high diversity in the current population, we execute a local search only for 10\% of all elite individuals (Line~\ref{alg:self_improvement_select_subset_elite_population}). Initially, we decode each individual to its phenotype consisting of a tour $\bpi$ and a packing plan $\bz$ (Line~\ref{alg:self_improvement_decode_p}). Then, the two phases of local optimizations are considered.
First, we apply a limited local search procedure (Line~\ref{alg:self_improvement_2opt_begin} to \ref{alg:self_improvement_2opt_end}) to the tour $\bpi$. The local search makes use of the well-known $2$-opt move, which has been successfully incorporated to solve various combinatorial optimization problems, including the single-objective TTP~\citep{el2016population, el2018efficiently}. We limited the number of $2$-opt moves to a small value $LS_{\bpi} = \min(100, n^2)$ since the exploitation phase may become computationally expensive when large instances are considered. After all $2$-opt moves have been executed, the tour $\bpi$ and the packing plan $\bz$ are added to the elite population $P_e$ if it is not dominated by any solution in $P_e$ (Line \ref{alg:self_improvement_update_nds}). Note that in the first phase, we do not change the packing plan $z$, which means the second objective of the problem (KP component) remains unchanged.
Second, we intend to improve the packing plan $\bz$ by applying bit-flip random moves (Line~\ref{alg:self_improvement_bit_flip_begin} to \ref{alg:self_improvement_bit_flip_end}). The bit-flip is also a well-known operator widely used in combinatorial optimization problems, including the single-objective TTP as well~\citep{faulkner2015approximate, chand2016fast}. Again because of the computational expensiveness, we apply $LS_{z} = \min(100, m)$ random bit-flip moves (Line~\ref{alg:self_improvement_select_random_bit_flip}-\ref{alg:self_improvement_performe_bit_flip}). As before, the new BI-TTP solution $(\pi, z')$ is insert into $P_e$ if it is not dominated by any solution in $P_e$ (Line~\ref{alg:self_improvement_update_nds2}). Finally, if $P_e$  contains more than $N_e$ solutions (Line~\ref{alg:self_improvement_ifcondition}), we select the best $N_e$ according to the their non-dominated rank and crowding distance (Line~\ref{alg:self_improvement_if}).

\begin{algorithm}
\makeatletter
\newcommand{\algorithmfootnote}[2][\footnotesize]{%
  \let\old@algocf@finish\@algocf@finish
  \def\@algocf@finish{\old@algocf@finish
    \leavevmode\rlap{\begin{minipage}{\linewidth}
    #1#2
    \end{minipage}}%
  }%
}
\DontPrintSemicolon
\SetKwData{Left}{left}
\SetKwData{Up}{up}
\SetKwFunction{FindCompress}{FindCompress}
\SetKwInOut{Input}{input}
\SetKwInOut{Output}{output}
\BlankLine
$\widehat{P_e} \gets $ select randomly $0.1N_e$ individuals from current elite population $P_e$ \label{alg:self_improvement_select_subset_elite_population} \\
\ForEach{$p \in \widehat{P_e}$} { 
    $(\pi, z) \gets$ decode $p$ in a feasible BI-TTP solution \label{alg:self_improvement_decode_p} \\
    \For{$i \gets 1$ \textbf{to} $LS_{\pi}$} { \label{alg:self_improvement_2opt_begin}
        $\pi' \gets$ generate a random 2-opt move in $\pi$ \\
        \lIf{\upshape $(\pi', z) \text{ is not dominated by } (\pi, z)$} {
            $(\pi, z) \gets (\pi', z)$
        }
    } \label{alg:self_improvement_2opt_end}
    \lIf{\upshape $(\pi, z) \text{ is not dominated by any solution in } P_e$} {
        $P_e \gets P_e \cup \{\;\zeta(\pi, z)\}$ \label{alg:self_improvement_update_nds}
    }
    \For{$i \gets 1$ \textbf{to} $LS_{z}$} {  \label{alg:self_improvement_bit_flip_begin}
        $item \gets $ select randomly an $item \in \{1, 2, \ldots, m\}$ \label{alg:self_improvement_select_random_bit_flip} \\
        \leIf{$item \in z$} { $z' \gets z \setminus \{item\}$ } { $z' \gets z \cup \{item\}$ } \label{alg:self_improvement_performe_bit_flip}
        \If{\upshape $(\pi, z') \text{ is not dominated by any solution in } P_e$} {
            $P_e \gets P_e \cup \{\;\zeta(\pi, z')\}$ \label{alg:self_improvement_update_nds2}
        }
    } \label{alg:self_improvement_bit_flip_end}
}
\If{$\vert P_e \vert \;> N_e$} { \label{alg:self_improvement_ifcondition}
    $P_e \gets $ select $N_{e}$ individuals according to the NSGA-II criteria \label{alg:self_improvement_if}
}
\caption{Exploitation phase.}
\label{alg:self_improvement}
\algorithmfootnote{$\zeta(\bpi, \bz)$ encodes the BI-TTP solution $(\bpi, \bz)$ to a vector of random-keys.}
\end{algorithm}%

To balance the exploration and exploitation phases in a run of the algorithm, we apply the exploitation phase only at every $\omega$ evolutionary cycles, which is another parameter of our proposed method.

\section{Computational Experiments}
\label{sec:computational-experiments}

In this section, we present the computational experiments we have employed to study the performance of our proposed method. We have chosen C/C++ as a programming language and have used BRKGA framework developed by \cite{toso2015c} and the Lin-Kernighan Heuristic (LKH), version 2.0.9\footnote{Available at \url{http://akira.ruc.dk/~keld/research/LKH/}}. The experiments have been executed on a high-performance cluster where each node is equipped with Intel(R) Xeon(R) 2.30 GHz processors. Each run of our algorithm has been sequentially (nonparallel) performed on a single processor. Our source code, as well as all non-dominated solutions found for each test instance, are available online\footnote{Available at \url{https://github.com/jonatasbcchagas/nds-brkga_bi-ttp}}.

In the following, we evaluate the performance of our proposed method on a variety of test instances. To be neither biased towards test instances with only a small or large number of cities and items, we have selected test instances with the purpose of covering different characteristics of the problem. Due to the design decisions we have made during the algorithm development, we provide a detailed parameter study to show the effectiveness of each customization of the evolutionary algorithm. Moreover, we present the rankings of competitions where we submitted our implementation to.

\subsection{Test instances}

In order to analyze the performance, we have considered nine medium/large instances from the comprehensive TTP benchmark developed by \cite{polyakovskiy2014comprehensive}. These instances, which are described in Table~\ref{table:instances}, have been used in the BI-TTP competitions at \textit{EMO-2019}\footnote{\url{https://www.egr.msu.edu/coinlab/blankjul/emo19-thief/}} and \textit{GECCO-2019}\footnote{\url{https://www.egr.msu.edu/coinlab/blankjul/gecco19-thief/}} conferences.

\begin{table*}
\footnotesize
\centering
\caption{BI-TTP test instances.}
\setlength{\tabcolsep}{0pt}
\begin{tabular*}{\hsize}{@{}@{\extracolsep{\fill}}llllll@{}}
\toprule
\multicolumn{1}{l}{\textbf{Instance}} & 
\multicolumn{1}{l}{\textbf{n}} & 
\multicolumn{1}{l}{\textbf{m}} & 
\multicolumn{1}{l}{\textbf{Q}} & 
\multicolumn{1}{l}{\textbf{Knapsack Type}} & \multicolumn{1}{l}{\textbf{R}} \\ 
\midrule
a280\_n279 & 280 & 279 & 25936 & bsc & 01 \\ 
a280\_n1395 &  & 1395 & 637010 & usw & 05 \\ 
a280\_n2790 &  & 2790 & 1262022 & unc & 10 \\
\midrule
fnl4461\_n4460 & 4461 & 4460 & 387150 & bsc & 01 \\ 
fnl4461\_n22300 &  & 22300 & 10182055 & usw & 05 \\ 
fnl4461\_n44600 &  & 44600 & 20244159 & unc & 10 \\
\midrule
pla33810\_n33809 & 33810 & 33809 & 2915215 & bsc & 01 \\ 
pla33810\_n169045 &  & 169045 & 77184794 & usw & 05 \\ 
pla33810\_n338090 &  & 338090 & 153960049 & unc & 10 \\ 
\bottomrule
\end{tabular*}
\label{table:instances}
\end{table*}

From Table~\ref{table:instances}, we can observe the characteristics of the instances, which involve 280 to 33810 cities (column $n$), 279 to 338090 items (column $m$), and knapsacks (column $Q$). Furthermore, the knapsack component of each instance has been built according to the profit/weight ratio of items in three different ways: bounded strongly correlated (bsc), uncorrelated with similar weights (usw), and uncorrelated (unc). To diversify the size of the knapsack component, it has been defined for each instance how many items per city are available (column $R$). For instance, when $R=10$, then $10$ items are available in each city. For example, assuming a problem with $280$ cities, this results in $2790$ items in total (as the first city never has any items).

\subsection{Parameter Study}

Customization often involves adding new parameters to the algorithm. Therefore, it is crucial to ensure the parameters are chosen well concerning the performance of the algorithm on a variety of test problems. 
For this reason, we investigate the influence of parameters in our proposed method. 
In the experiment, we run a systemic setup of parameters to finally draw conclusions regarding their performance on the different types of test instances. Finally, we provide suggestions on how to choose parameters for new unknown problems.

Our proposed method has eight parameters in total which are shown in Table~\ref{tbl:parameters-overview}: Population size $N$, elite population size $N_{e}$, mutant population size $N_{m}$, elite allele inheritance probability $\rho_{e}$, fraction $\alpha$ of the initial population created from TSP and KP solutions (see Algorithm~\ref{alg:initial_population}), and the frequency $\omega$, in terms of evolutionary cycles in which a local search is applied (see Algorithm~\ref{alg:self_improvement}). Furthermore, two subproblem dependent parameters have to be defined: $t$, which is the upper bound for the time of the TSP solvers to be executed, and $\delta$, which is the number of different KP capacities that should be considered.
To evaluate the influence of each parameter, we conduct several experiments.

\begin{table}
\footnotesize
\centering
\caption{Parameter overview.}
\begin{tabular}{p{1.5cm}p{6cm}}
\toprule
\textbf{Parameter} & \textbf{Description}\\ 
\midrule
$N$ & Population Size \\[1mm]
$N_e$ & Number of Elites \\[1mm]
$N_m$ & Number of Mutant Individuals \\[1mm]
$\rho_e$ & Probability of Elite during Biased Crossover \\[1mm]
$\alpha$ & Fraction of near-optimal Solutions obtained by solving TSP or KP independently\\[1mm]
$\omega$ &  Frequency of Local Search Procedures \\[1.5mm]
TSP - t &  Time in Seconds to solve the TSP problem \\[1.5mm]
KP - $\delta$ & Gap of Knapsack Capacity $Q$ in between different KP Optimizer Runs \\[1mm]
\bottomrule
\end{tabular}
\label{tbl:parameters-overview}
\end{table}

In this parameter study, we first investigate the effect of $t$ and $\delta$ on the performance. Both variables affect the initial population that consists partly of solutions from TSP and KP solvers. For solving the TSP component independently, we have used the Lin-Kernighan Heuristic (LKH). The LKH is one of the most efficient algorithms for generating optimal or near-optimal solutions for the symmetric traveling salesman problem. Naturally, the LKH has higher computational costs as TSP instances increase. To balance the computational cost and the quality of the solution achieved, we limit the LKH execution time to different values and compare the obtained solution with the optimal solution. As LKH has random components, we run it ten independent times and use the average reached by them. In Table \ref{table:lkh_time_influence}, for each TSP instance and each runtime, we show the relative percentage difference between the solution obtained with limited time and the TSP optimal solution.

\begin{table}
\footnotesize
\centering
\caption{Influence of execution time on the LKH heuristic algorithm.}
\setlength{\tabcolsep}{0pt}
\begin{tabular*}{\hsize}{@{}@{\extracolsep{\fill}}llrrrrrrr@{}}
\toprule
\multicolumn{1}{c}{\textbf{TSP}} &  & \multicolumn{7}{l}{t in seconds} \\
\cmidrule{3-9}
\multicolumn{1}{c}{comp.} &  & \multicolumn{1}{l}{60} & \multicolumn{1}{l}{180} & \multicolumn{1}{l}{300} & \multicolumn{1}{l}{420} & \multicolumn{1}{l}{600} & \multicolumn{1}{l}{1800} & \multicolumn{1}{l}{3600} \\ 
\midrule
a280 &  & 0.0000\% & 0.0000\% & 0.0000\% & 0.0000\% & 0.0000\% & 0.0000\% & 0.0000\% \\ 
fnl4461 &  & 0.0180\% & 0.0078\% & 0.0035\% & 0.0028\% & 0.0028\% & 0.0000\% & 0.0000\% \\ 
pla33810 &  & 0.6653\% & 0.5964\% & 0.4261\% & 0.3339\% & 0.2377\% & 0.1084\% & 0.0837\% \\
\midrule
Avg. &  & 0.2278\% & 0.2014\% & 0.1432\% & 0.1122\% & 0.0802\% & 0.0361\% & 0.0279\% \\
\bottomrule
\end{tabular*}
\label{table:lkh_time_influence}
\end{table}

Table \ref{table:lkh_time_influence} shows that even for shorter computational times LKH is efficient. On average, LKH has been able to find solutions with a gap of less than $0.23\%$ to the optimal solutions, even considering only $60$ seconds of processing. Naturally, the quality of solutions increases with longer computational time. As we do not pursue spending a significant amount of time solving the TSP independently, we have limited the LKH to $300$ seconds in our implementation. The experiment indicates that this is sufficient to produce near-optimal TSP solutions to be used in the initial population.\footnote{Note that we rotate the computed tours to conform with the requirement for all TTP tours to start and finish in city number 1.}

Moreover, the parameter $\delta$ used in the KP solver has to be studied and the performance of the GH+DP algorithm evaluated. This algorithm has been developed for solving the KP component independently. For each KP instance, we have run the GH+DP for different values of $\delta$ and have measured the quality of the solutions obtained. Table~\ref{table:ghdp_delta_influence} shows the difference between the solution obtained and the KP optimal solution $d_{p}^{*}$ for different $\delta$ values. In addition to the difference, it provides the computational time required in seconds $t(s)$. We can observe that for almost all instances, GH+DP has been able to find the KP optimal solution even with a relatively small $\delta$. Moreover, the larger $\delta$, the larger the fraction of the knapsack solved using the dynamic programming algorithm, hence the higher quality of the solution and the longer the computation time. In order to find the best possible solutions for the KP component within a reasonable time, we have chosen to use $\delta = 5\cdot10^4$.

\begin{table}
\footnotesize
\centering
\caption{Influence of $\delta$ on GH+DP algorithm.}
\setlength{\tabcolsep}{0pt}
\begin{tabular*}{\hsize}{@{}@{\extracolsep{\fill}}llrrlrrlrrlrrlrrlrr@{}}
\toprule
\multicolumn{1}{l}{\multirow{2}{*}{KP comp.}} & & \multicolumn{2}{c}{$\delta = 10^3$} & & \multicolumn{2}{c}{$\delta = 5 \cdot 10^3$} & & \multicolumn{2}{c}{$\delta = 10^4$} & & \multicolumn{2}{c}{$\delta = 5 \cdot 10^4$} & & \multicolumn{2}{c}{$\delta = 10^5$} & & \multicolumn{2}{c}{$\delta = 5 \cdot 10^5$} \\ 
\cmidrule{3-4} \cmidrule{6-7} \cmidrule{9-10} \cmidrule{12-13} \cmidrule{15-16} \cmidrule{18-19}  
\multicolumn{ 1}{l}{} & & \multicolumn{ 1}{c}{$d_{p}^{*}$} & \multicolumn{ 1}{c}{t(s)} & & \multicolumn{ 1}{c}{$d_{p}^{*}$} & \multicolumn{ 1}{c}{t(s)} & & \multicolumn{ 1}{c}{$d_{p}^{*}$} & \multicolumn{ 1}{c}{t(s)} & & \multicolumn{ 1}{c}{$d_{p}^{*}$} & \multicolumn{ 1}{c}{t(s)} & & \multicolumn{ 1}{c}{$d_{p}^{*}$} & \multicolumn{ 1}{c}{t(s)} & & \multicolumn{ 1}{c}{$d_{p}^{*}$} & \multicolumn{ 1}{c}{t(s)} \\ 
\midrule
n279 &  & 0 & 0 &  & 0 & 0 &  & 0 & 0 &  & 0 & 0 &  & 0 & 0 &  & 0 & 2\\ 
n1395 &  & 0 & 0 &  & 0 & 0 &  & 0 & 0 &  & 0 & 3 &  & 0 & 33 &  & 0 & 1364 \\ 
n2790 &  & 0 & 0 &  & 0 & 0 &  & 0 & 0 &  & 0 & 2 &  & 0 & 18 &  & 0 & 1339 \\[1mm] 
n4460 &  & 0 & 0 &  & 0 & 0 &  & 0 & 1 &  & 0 & 54 &  & 0 & 162 &  & 0 & 9007 \\ 
n22300 &  & 27 & 0 &  & 12 & 1 &  & 12 & 2 &  & 0 & 105 &  & 0 & 265 &  & 0 & 10234 \\ 
n44600 &  & 0 & 0 &  & 0 & 0 &  & 0 & 1 &  & 0 & 47 &  & 0 & 136 &  & 0 & 5193 \\[1mm] 
n33809 &  & 0 & 1 &  & 0 & 3 &  & 0 & 6 &  & 0 & 261 &  & 0 & 1149 &  & 0 & 22945 \\ 
n169045 &  & 298 & 3 &  & 298 & 7 &  & 295 & 14 &  & 228 & 437 &  & 11 & 3980 &  & 0 & 35711 \\ 
n338090 &  & 0 & 1 &  & 0 & 2 &  & 0 & 5 &  & 0 & 226 &  & 0 & 928 &  & 0 & 19594 \\ 
\midrule
Avg. & & 36.1 & 0.6 &  & 34.4 & 1.4 &  & 34.1 & 3.2 &  & 32.0 & 126.1 &  & 1.2 & 741.2 &  & 0.0 & 11708.8 \\
\bottomrule
\end{tabular*}
\label{table:ghdp_delta_influence}
\end{table}

This preliminary study on subproblem solvers has shown that $t=300$ for the TSP solver and $\delta = 5\cdot10^4$ for KP solver seem to be reasonable parameters regarding the trade-off of running time and quality of solutions. So far, we have evaluated the goodness of each of the subproblems without considering the interwovenness aspect. Next, the parameter $\alpha$ defines how many solutions are used from those subproblem solvers during the initialization.
To draw conclusions about the remaining six parameters, we have conducted an experiment with predefined parameter settings.
The considered parameter values for each parameter are shown in Table~\ref{table:ndsbrkga-parameters}. We have considered all $3072$ possible combinations that can be formed by combining these values.
Because the proposed method contains components with underlying randomness, we have run each parameter configuration $10$ times for $5$ hours. Altogether, the experiments have consumed $1,382,400$ CPU hours, which is equivalent to almost $158$ CPU years.

\begin{table}
\footnotesize
\centering
\caption{Parameter values considered during the experiment.}
\begin{tabular}{p{1.5cm}p{4cm}}
\toprule
\textbf{Parameter} & \textbf{Values} \\ 
\midrule
$N$ & 100, 200, 500, 1000 \\[1mm]
$N_e$ & 0.3$N$, 0.4$N$, 0.5$N$, 0.6$N$ \\[1mm]
$N_m$ & 0.0$N$, 0.1$N$, 0.2$N$ \\[1mm]
$\rho_e$ & 0.5, 0.6, 0.7, 0.8 \\[1mm]
$\alpha$ & 0.0, 0.1, 0.2, 0.3 \\[1mm]
$\omega$ & 1, 10, 50, 100 \\[1mm]
\bottomrule
\end{tabular}
\label{table:ndsbrkga-parameters}
\end{table}

We use the hypervolume indicator (HV)~\citep{zitzler1998multiobjective} as a performance indicator to compare and analyze results obtained from the set of parameter configurations. It is one of the most used indicators for measuring the quality of a set of non-dominated solutions by calculating the volume of the dominated portion of the objective space bounded from a reference point. Considering the BI-TTP, it considers the dominated volume regarding the minimum time and the maximum profit. Note that maximizing the hypervolume indicator is equivalent to finding a good approximation of the Pareto front,\footnote{However, maximizing the hypervolume is not equivalent to finding the optimal approximation, see, e.g.,~\cite{BRINGMANN2013265,WAGNER2015465}} thereby the higher the hypervolume indicator, the better the solution sets are (in general terms). To make the hypervolume suitable for the comparison of objectives with greatly varying ranges, these need to be normalized beforehand. Therefore, we have first normalized the values of the objectives between 0 and 1 according to their minimum and maximum value found during the parameter tuning experiments before computing the hypervolume.

\begin{figure*}[!ht]
    \captionsetup[subfigure]{justification=centering, labelformat=empty}
    \centering
    \subfloat[a280\_n279]{\includegraphics[width=0.33\textwidth]{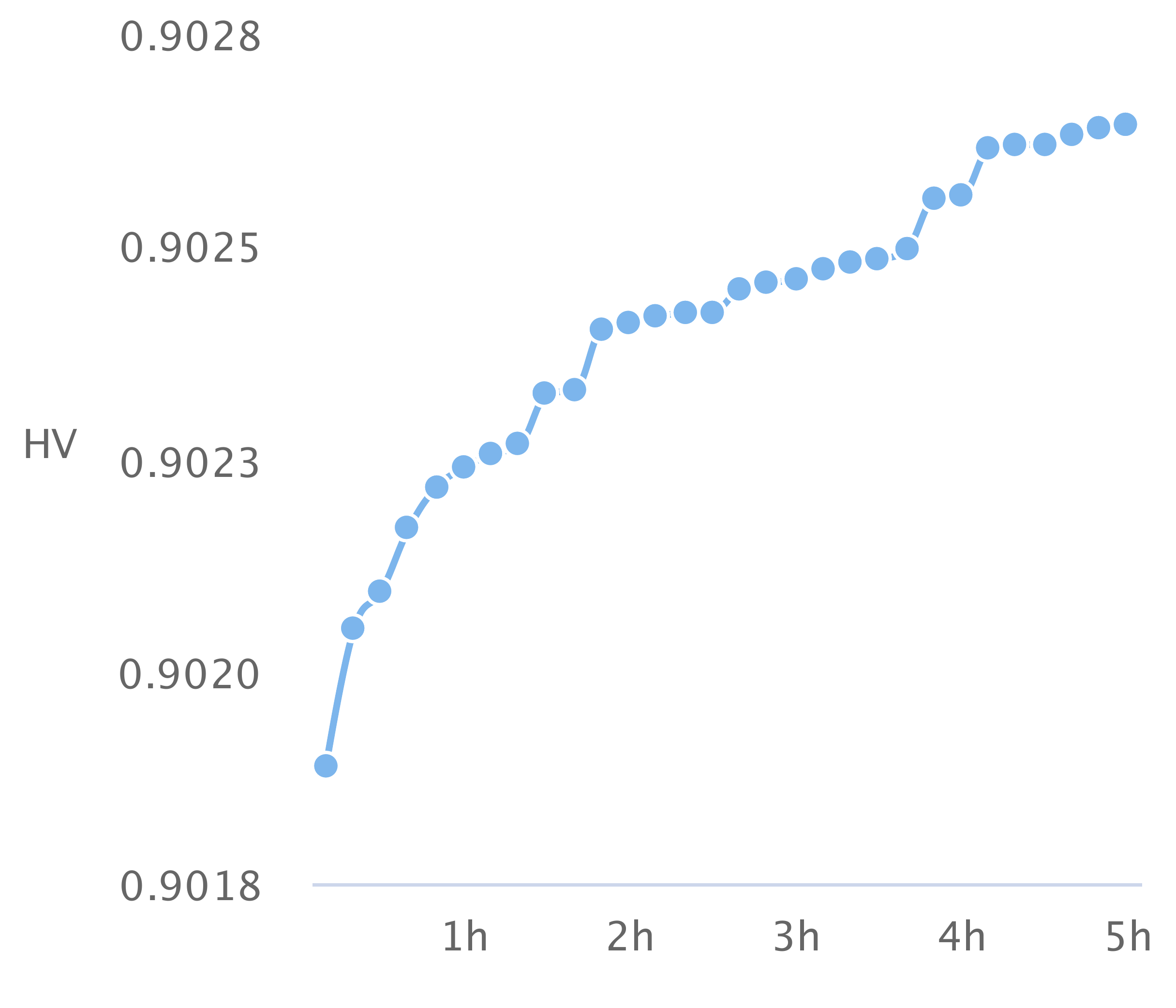}}
    \subfloat[a280\_n1395]{\includegraphics[width=0.33\textwidth]{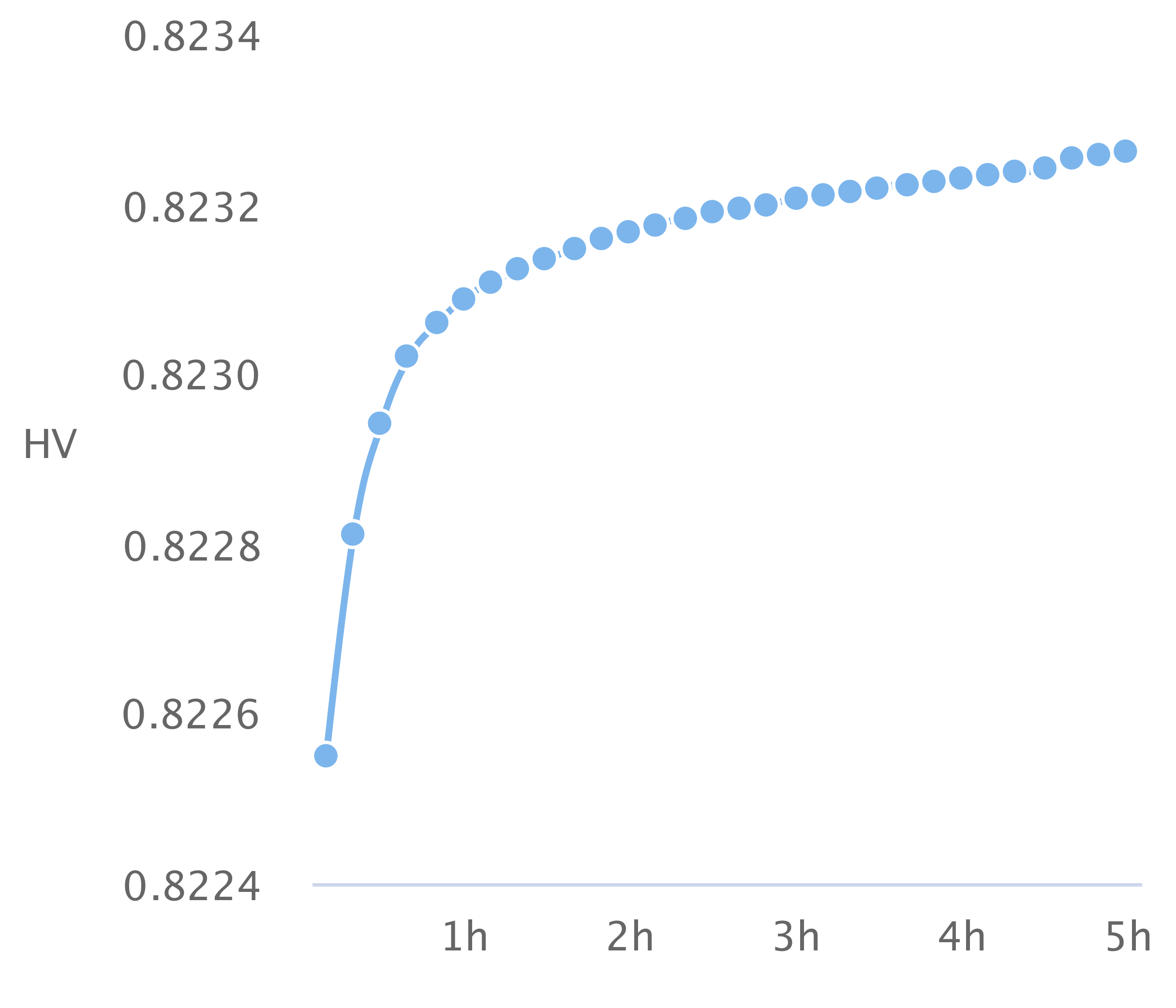}}
    \subfloat[a280\_n2790]{\includegraphics[width=0.33\textwidth]{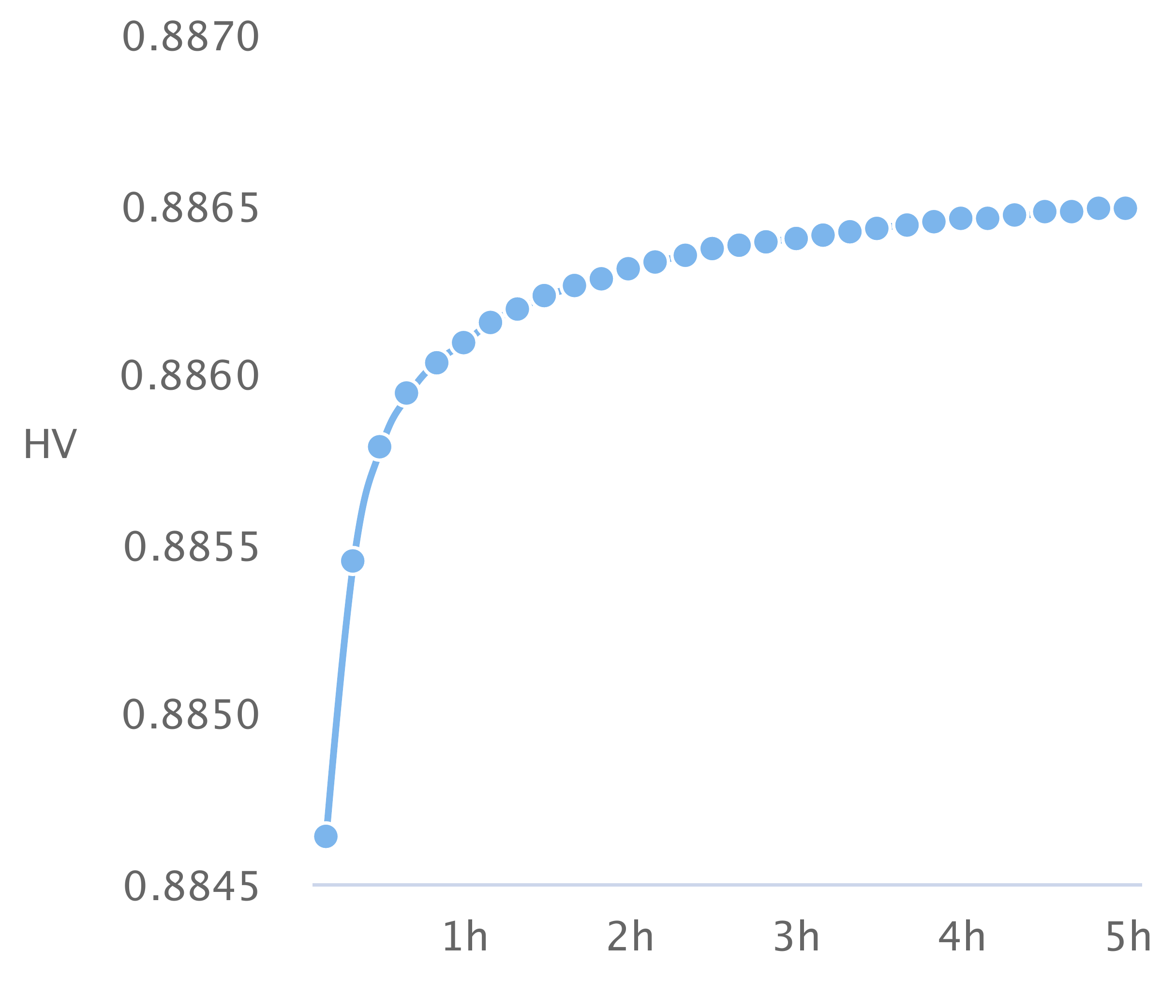}}
    
    \subfloat[fnl4461\_n4460]{\includegraphics[width=0.33\textwidth]{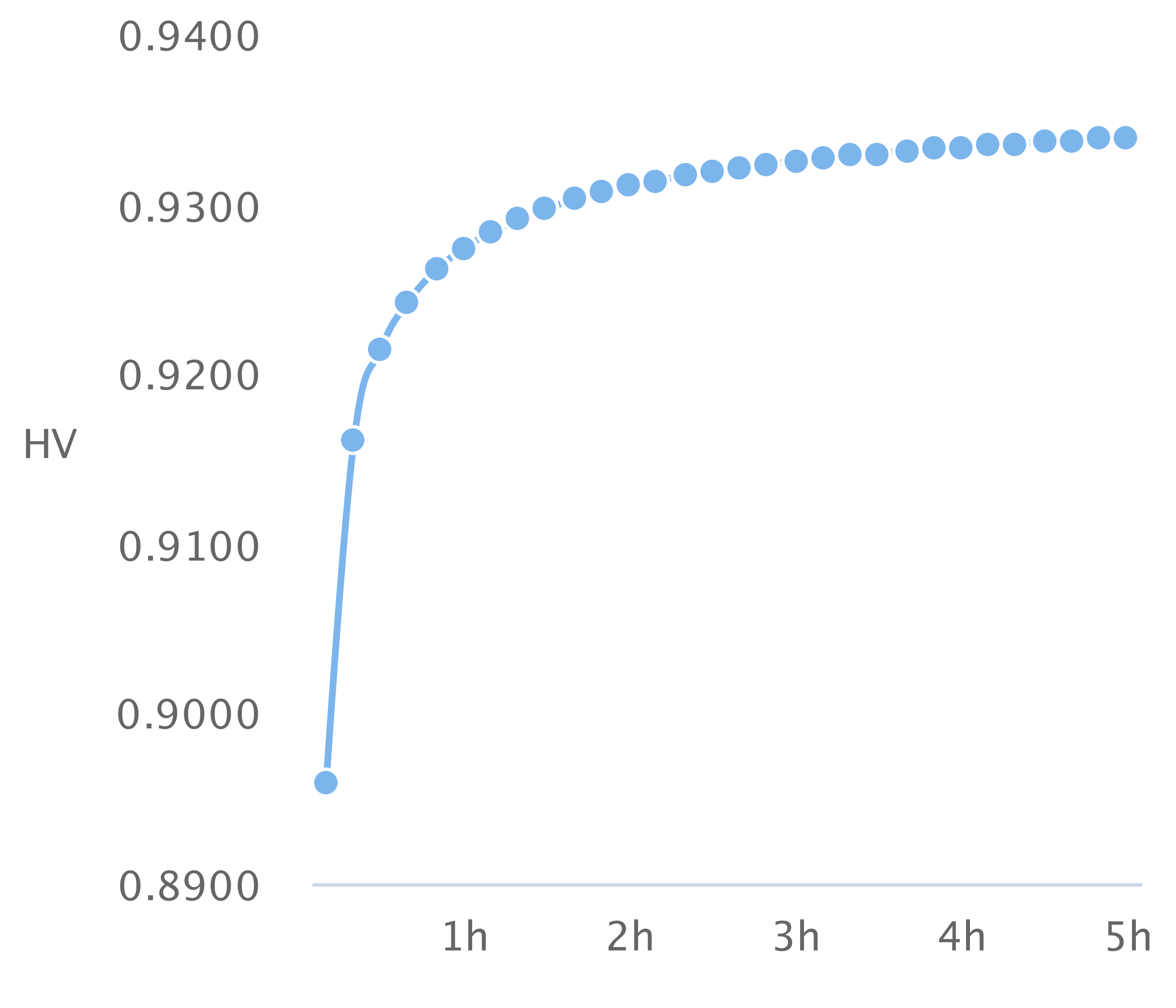}}
    \subfloat[fnl4461\_n22300]{\includegraphics[width=0.33\textwidth]{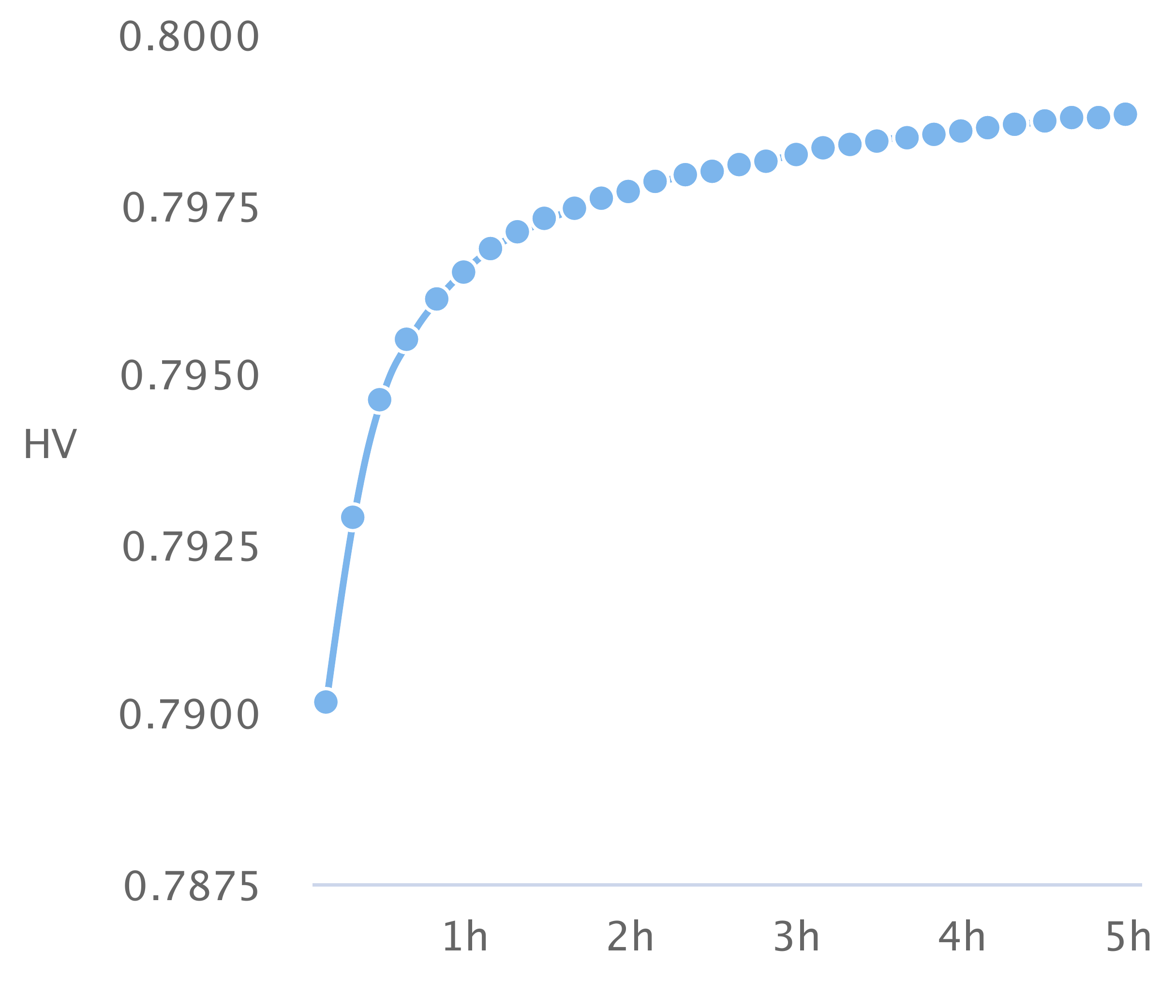}}
    \subfloat[fnl4461\_n44600]{\includegraphics[width=0.33\textwidth]{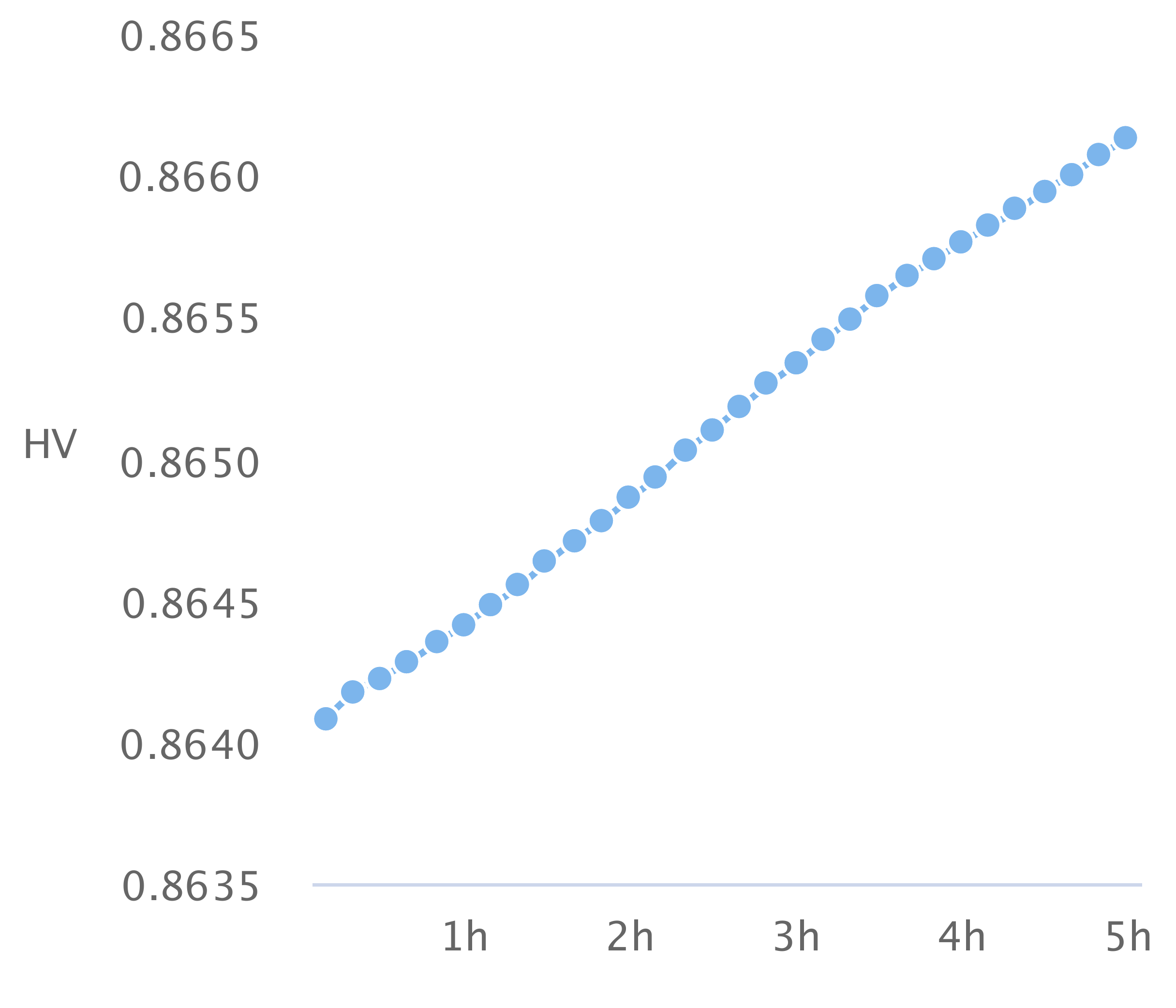}}
    
    \subfloat[pla33810\_n33809]{\includegraphics[width=0.33\textwidth]{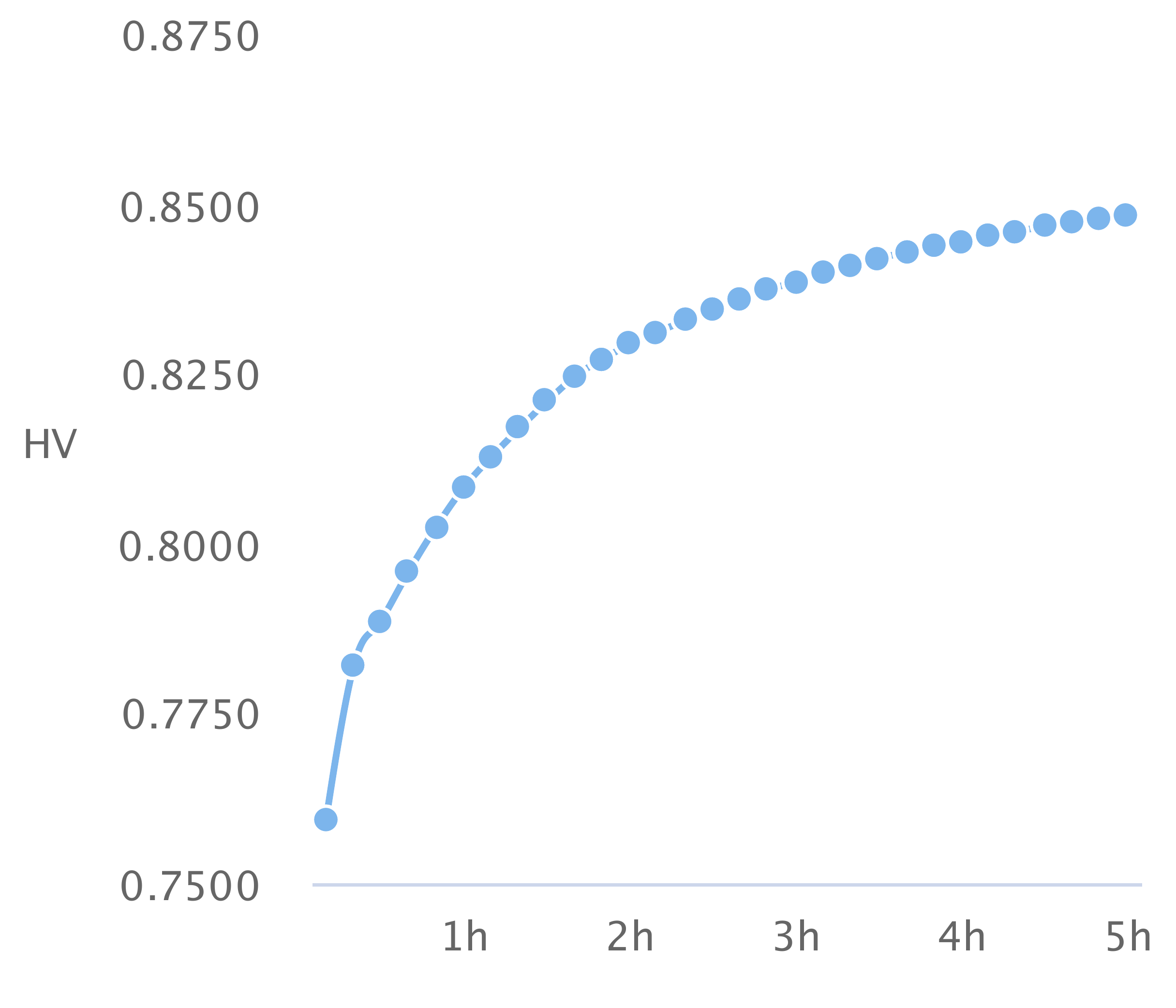}}
    \subfloat[pla33810\_n169045]{\includegraphics[width=0.33\textwidth]{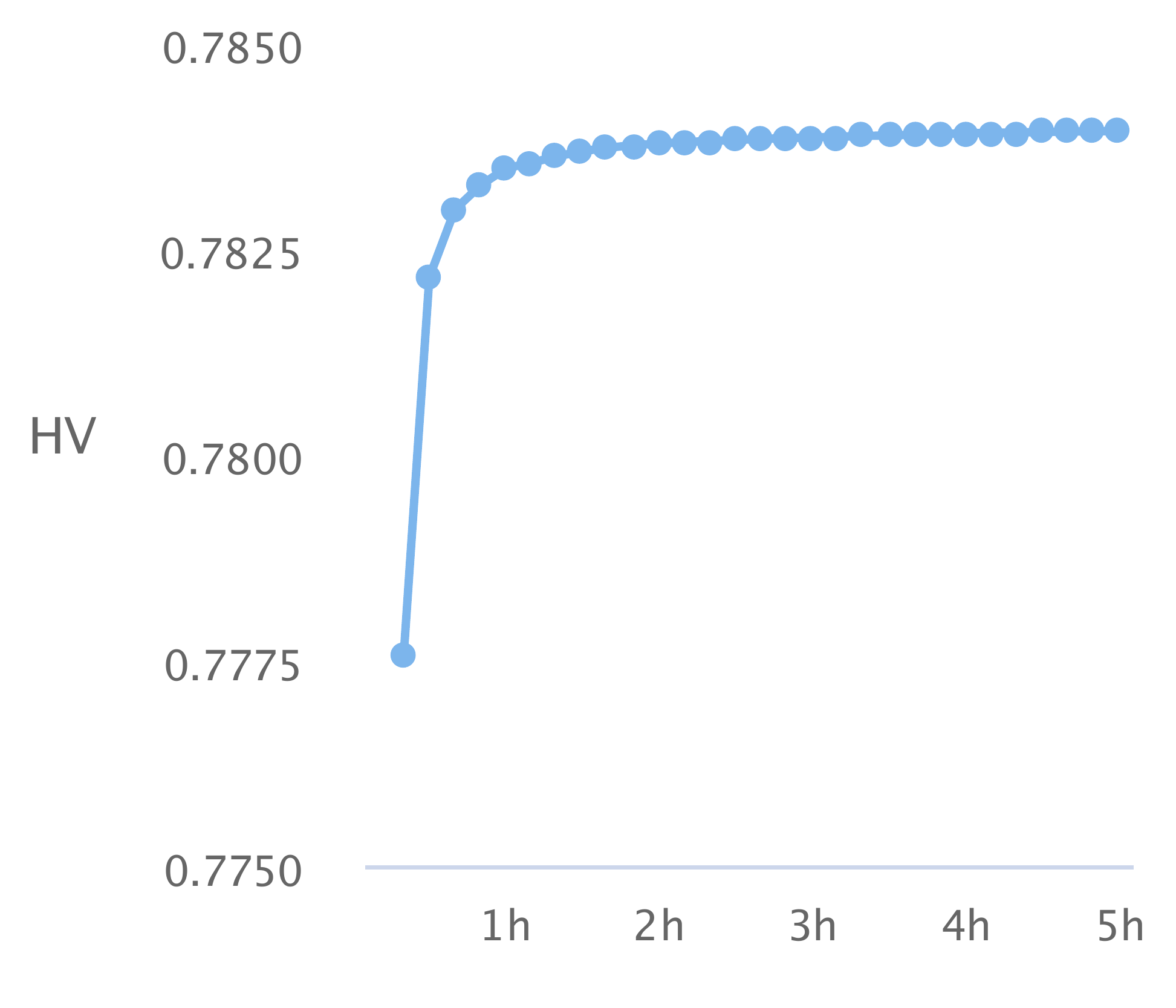}}
    \subfloat[pla33810\_n338090]{\includegraphics[width=0.33\textwidth]{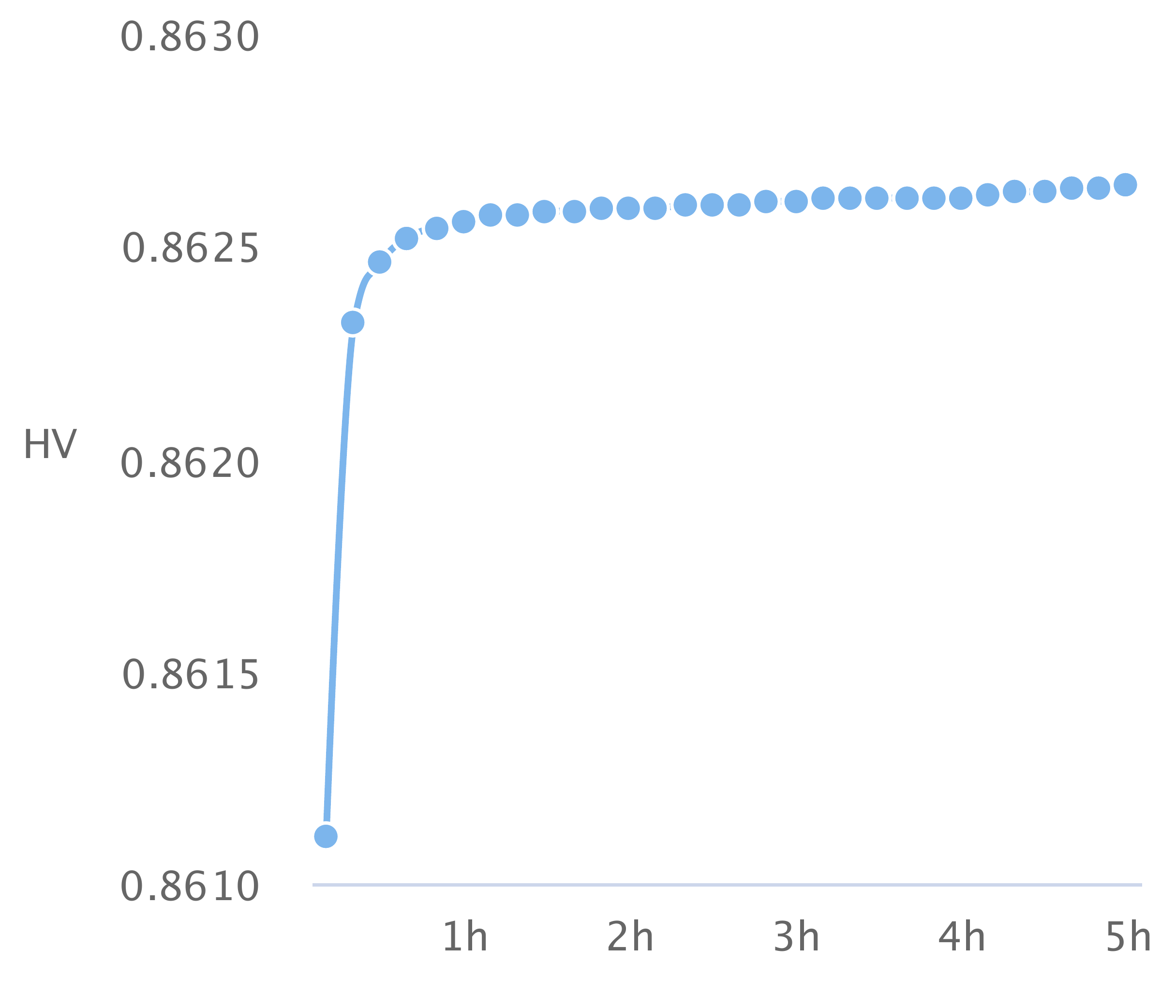}}
    \caption{Convergence plots according to the hypervolume indicator.}
    \label{fig:convergence_plots_hv}
\end{figure*}

Figure \ref{fig:convergence_plots_hv} shows the convergence according to the hypervolume indicator for each instance throughout $5$ hours, considering $10$-minute intervals. For each interval, we have plotted the result of the best parameter configuration found. Each parameter configuration is described in Table \ref{table:convergence_hv}. It is important to note that the vertical axis (hypervolume values) of the plots are not on the same scale. The figure shows that our proposed method has been able to quickly converge for most of the instances, indicating that our algorithm does not need excessive processing time to achieve good solutions. 

\begin{table}
\footnotesize
\centering
\caption{Best parameter configurations as observed in the parameter study.}
\begin{tabular*}{\hsize}{@{}@{\extracolsep{\fill}}lcrrrrrr@{}}
\toprule
\multicolumn{ 1}{c}{Instance} & \multicolumn{ 1}{c}{Runtime} & \multicolumn{ 6}{c}{Best parameter configuration} \\ 
\cmidrule{3-8}
\multicolumn{ 1}{l}{} & \multicolumn{ 1}{c}{} & \multicolumn{1}{c}{$N$} & \multicolumn{1}{c}{$N_e$} & \multicolumn{1}{c}{$N_m$} & \multicolumn{1}{c}{$\rho_{e}$} & \multicolumn{1}{c}{$\alpha$} & \multicolumn{1}{c}{$\omega$} \\ 
\midrule
a280\_n279 & 600 & 500 & 0.4 & 0.1 & 0.7 & 0.1 & 10 \\ 
 & 1200 & 500 & 0.4 & 0.1 & 0.7 & 0.2 & 10 \\ 
 & 1800 & 500 & 0.5 & 0.0 & 0.5 & 0.2 & 10 \\ 
 & 2400 - 18000 & 500 & 0.3 & 0.2 & 0.8 & 0.3 & 50 \\ 
\midrule
a280\_n1395 & 600 & 1000 & 0.6 & 0.0 & 0.6 & 0.3 & 10 \\ 
 & 1200 & 1000 & 0.6 & 0.0 & 0.5 & 0.2 & 10 \\ 
 & 1800 - 2400 & 1000 & 0.6 & 0.1 & 0.5 & 0.2 & 50 \\ 
 & 3000 - 4800 & 1000 & 0.6 & 0.0 & 0.5 & 0.2 & 50 \\ 
 & 5400 - 9600 & 1000 & 0.6 & 0.0 & 0.6 & 0.2 & 50 \\ 
 & 10200 - 15600 & 1000 & 0.6 & 0.0 & 0.5 & 0.3 & 100 \\ 
 & 16200 - 18000 & 1000 & 0.6 & 0.1 & 0.8 & 0.3 & 50 \\ 
\midrule
a280\_n2790 & 600 & 1000 & 0.6 & 0.0 & 0.6 & 0.1 & 100 \\ 
 & 1200 & 1000 & 0.6 & 0.0 & 0.5 & 0.1 & 50 \\ 
 & 1800 - 3600 & 1000 & 0.6 & 0.0 & 0.7 & 0.2 & 50 \\ 
 & 4200 & 1000 & 0.6 & 0.0 & 0.7 & 0.1 & 50 \\ 
 & 4800 - 6000 & 1000 & 0.6 & 0.0 & 0.5 & 0.3 & 50 \\ 
 & 6600 - 12000 & 1000 & 0.6 & 0.0 & 0.5 & 0.1 & 50 \\ 
 & 12600 - 14400 & 1000 & 0.6 & 0.0 & 0.5 & 0.3 & 50 \\ 
 & 15000 & 1000 & 0.6 & 0.0 & 0.5 & 0.2 & 50 \\ 
 & 15600 - 18000 & 1000 & 0.6 & 0.0 & 0.5 & 0.3 & 50 \\ 
\midrule
fnl4461\_n4460 & 600 & 500 & 0.6 & 0.0 & 0.5 & 0.1 & 50 \\ 
 & 1200 - 4200 & 1000 & 0.6 & 0.0 & 0.7 & 0.1 & 100 \\ 
 & 4800 - 9600 & 1000 & 0.6 & 0.0 & 0.6 & 0.2 & 100 \\ 
 & 10200 - 13800 & 1000 & 0.6 & 0.0 & 0.5 & 0.2 & 50 \\ 
 & 14400 - 18000 & 1000 & 0.6 & 0.0 & 0.5 & 0.3 & 50 \\ 
\midrule
fnl4461\_n22300 & 600 & 1000 & 0.5 & 0.0 & 0.8 & 0.1 & 10 \\ 
 & 1200 & 1000 & 0.6 & 0.0 & 0.8 & 0.2 & 50 \\ 
 & 1800 - 6600 & 1000 & 0.6 & 0.0 & 0.5 & 0.2 & 100 \\ 
 & 7200 & 1000 & 0.6 & 0.0 & 0.5 & 0.3 & 100 \\ 
 & 7800 & 1000 & 0.6 & 0.0 & 0.5 & 0.2 & 100 \\ 
 & 8400 - 10200 & 1000 & 0.6 & 0.0 & 0.5 & 0.3 & 100 \\ 
 & 10800 - 18000 & 1000 & 0.6 & 0.0 & 0.5 & 0.1 & 50 \\ 
\midrule
fnl4461\_n44600 & 600 & 1000 & 0.6 & 0.0 & 0.8 & 0.3 & 100 \\ 
 & 1200 & 1000 & 0.6 & 0.0 & 0.8 & 0.3 & 100 \\ 
 & 1800 - 15600 & 1000 & 0.6 & 0.0 & 0.5 & 0.3 & 100 \\ 
 & 16200 - 18000 & 1000 & 0.6 & 0.0 & 0.5 & 0.2 & 100 \\ 
\midrule
pla33810\_n33809 & 600 & 1000 & 0.3 & 0.2 & 0.5 & 0.1 & 1 \\ 
 & 1200 & 1000 & 0.6 & 0.0 & 0.7 & 0.2 & 10 \\ 
 & 1800 & 1000 & 0.5 & 0.0 & 0.8 & 0.3 & 10 \\ 
 & 2400 & 1000 & 0.6 & 0.0 & 0.7 & 0.3 & 50 \\ 
 & 3000 & 1000 & 0.6 & 0.0 & 0.5 & 0.2 & 50 \\ 
 & 3600 & 1000 & 0.6 & 0.0 & 0.6 & 0.2 & 50 \\ 
 & 4200 & 1000 & 0.6 & 0.0 & 0.5 & 0.2 & 50 \\ 
 & 4800 & 1000 & 0.6 & 0.0 & 0.6 & 0.1 & 50 \\ 
 & 5400 & 1000 & 0.6 & 0.0 & 0.6 & 0.3 & 50 \\ 
 & 6000 - 11400 & 1000 & 0.6 & 0.0 & 0.6 & 0.1 & 50 \\ 
 & 12000 - 18000 & 1000 & 0.6 & 0.0 & 0.6 & 0.1 & 10 \\ 
\midrule
pla33810\_n169045 & 600 & 500 & 0.5 & 0.0 & 0.8 & 0.3 & 10 \\ 
 & 1200 & 100 & 0.3 & 0.1 & 0.8 & 0.3 & 1 \\ 
 & 1800 & 500 & 0.5 & 0.0 & 0.8 & 0.2 & 1 \\ 
 & 2400 - 3000 & 500 & 0.3 & 0.0 & 0.8 & 0.3 & 1 \\ 
 & 3600 - 13800 & 1000 & 0.3 & 0.0 & 0.8 & 0.3 & 1 \\ 
 & 14400 - 18000 & 1000 & 0.4 & 0.2 & 0.8 & 0.3 & 1 \\ 
\midrule
pla33810\_n338090 & 600 & 1000 & 0.3 & 0.1 & 0.8 & 0.3 & 1 \\ 
 & 1200 - 4800 & 1000 & 0.6 & 0.0 & 0.8 & 0.3 & 100 \\ 
 & 5400 - 8400 & 1000 & 0.6 & 0.0 & 0.7 & 0.3 & 100 \\ 
 & 9000 - 14400 & 1000 & 0.6 & 0.2 & 0.5 & 0.3 & 100 \\ 
 & 15000 & 1000 & 0.4 & 0.0 & 0.8 & 0.3 & 1 \\ 
 & 15600 - 18000 & 1000 & 0.3 & 0.1 & 0.8 & 0.3 & 1 \\ 
 \bottomrule
\end{tabular*}
\label{table:convergence_hv}
\end{table}

Table \ref{table:convergence_hv} shows that the best parameter configuration for each instance changes over runtime. However, the number of changes among them decreases as the runtime increases, which means our method keeps stable regarding its best parameter configuration as the runtime increases. Although a single parameter configuration cannot extract the best performance by considering all instances, we can observe some patterns and trends among the different parameter configurations. 
A good parameter configuration is a population with a larger number of individuals, higher survival rate for the best individuals, insignificant contribution from mutant individuals, and high contribution of the TSP and KP solvers for creating part of the initial population.

\begin{figure*}[!ht]
    \captionsetup[subfigure]{justification=centering, labelformat=empty}
    \centering
    \subfloat[600 seconds]{\includegraphics[width=0.45\textwidth]{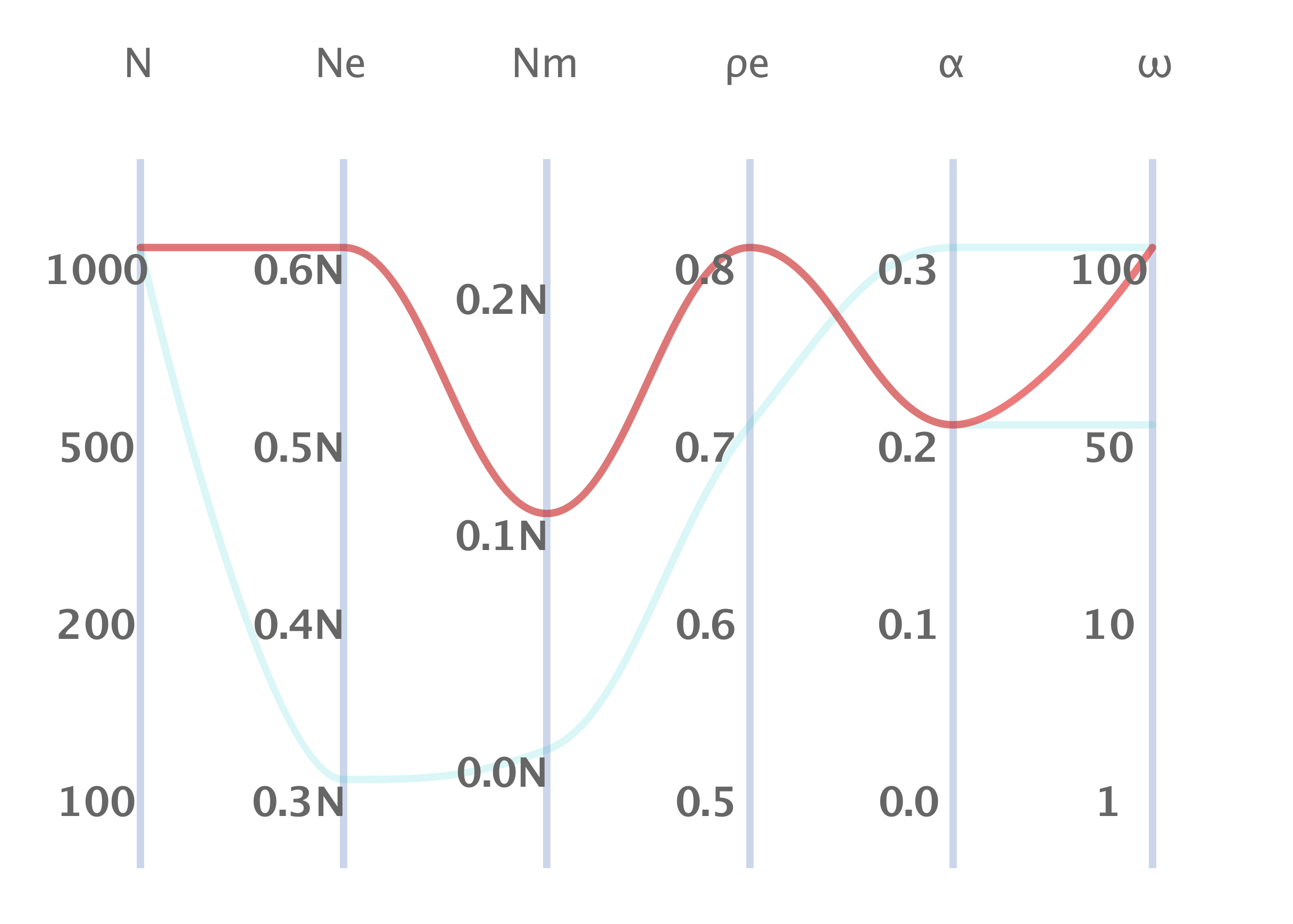}}
    \qquad \quad
    \subfloat[1200 seconds]{\includegraphics[width=0.45\textwidth]{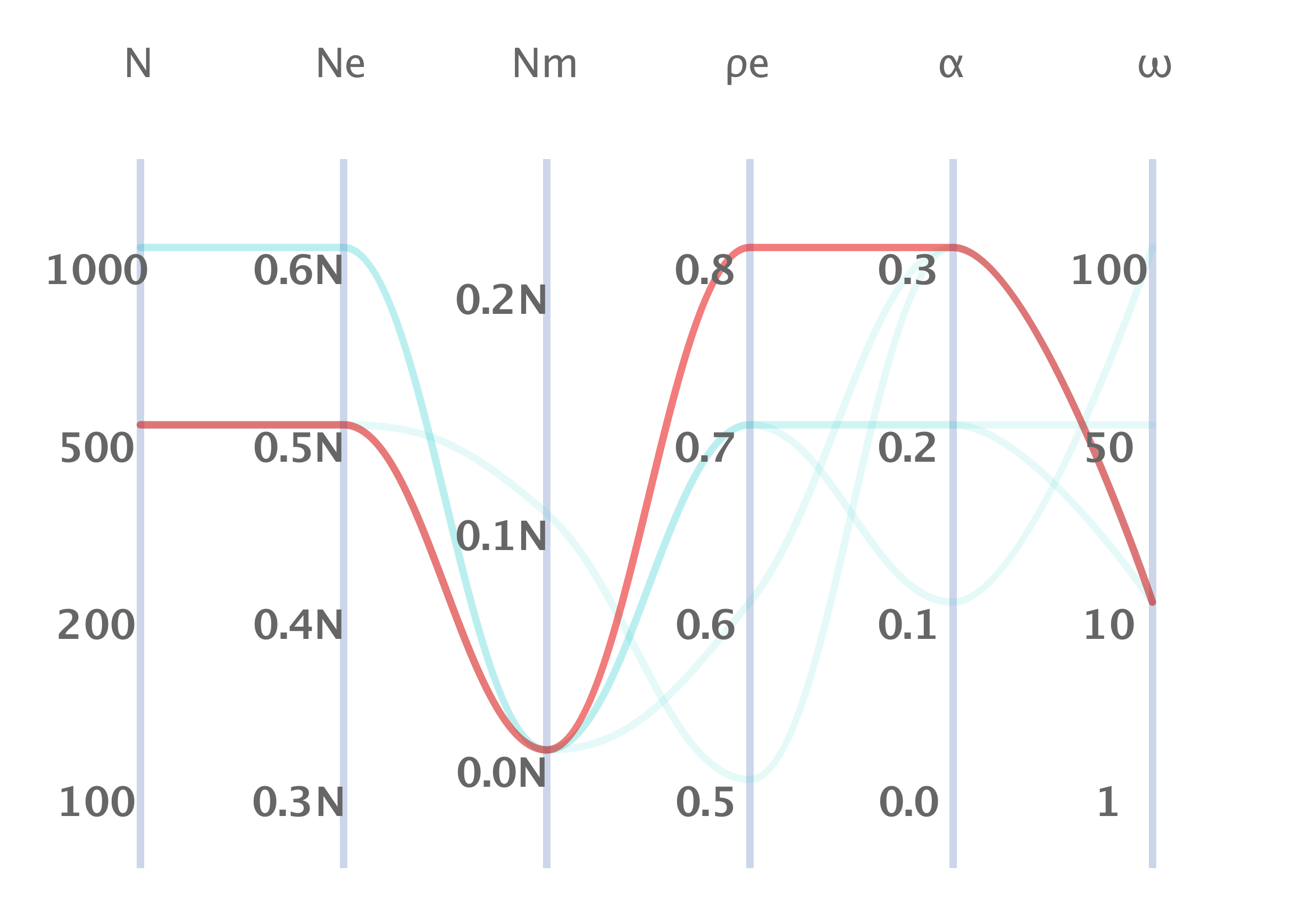}}
    
    \subfloat[1800 seconds]{\includegraphics[width=0.45\textwidth]{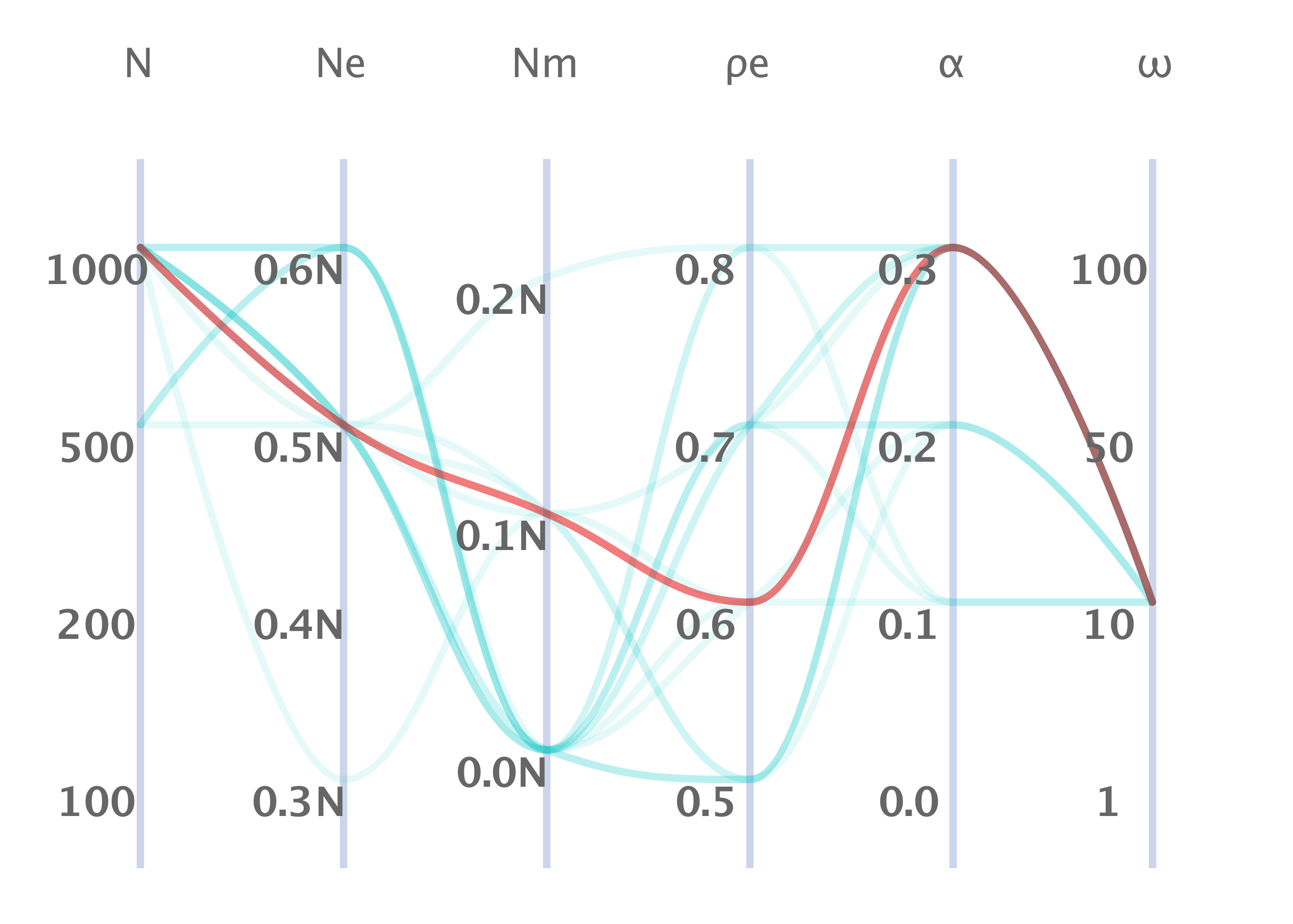}}    
    \qquad \quad
    \subfloat[3600 seconds]{\includegraphics[width=0.45\textwidth]{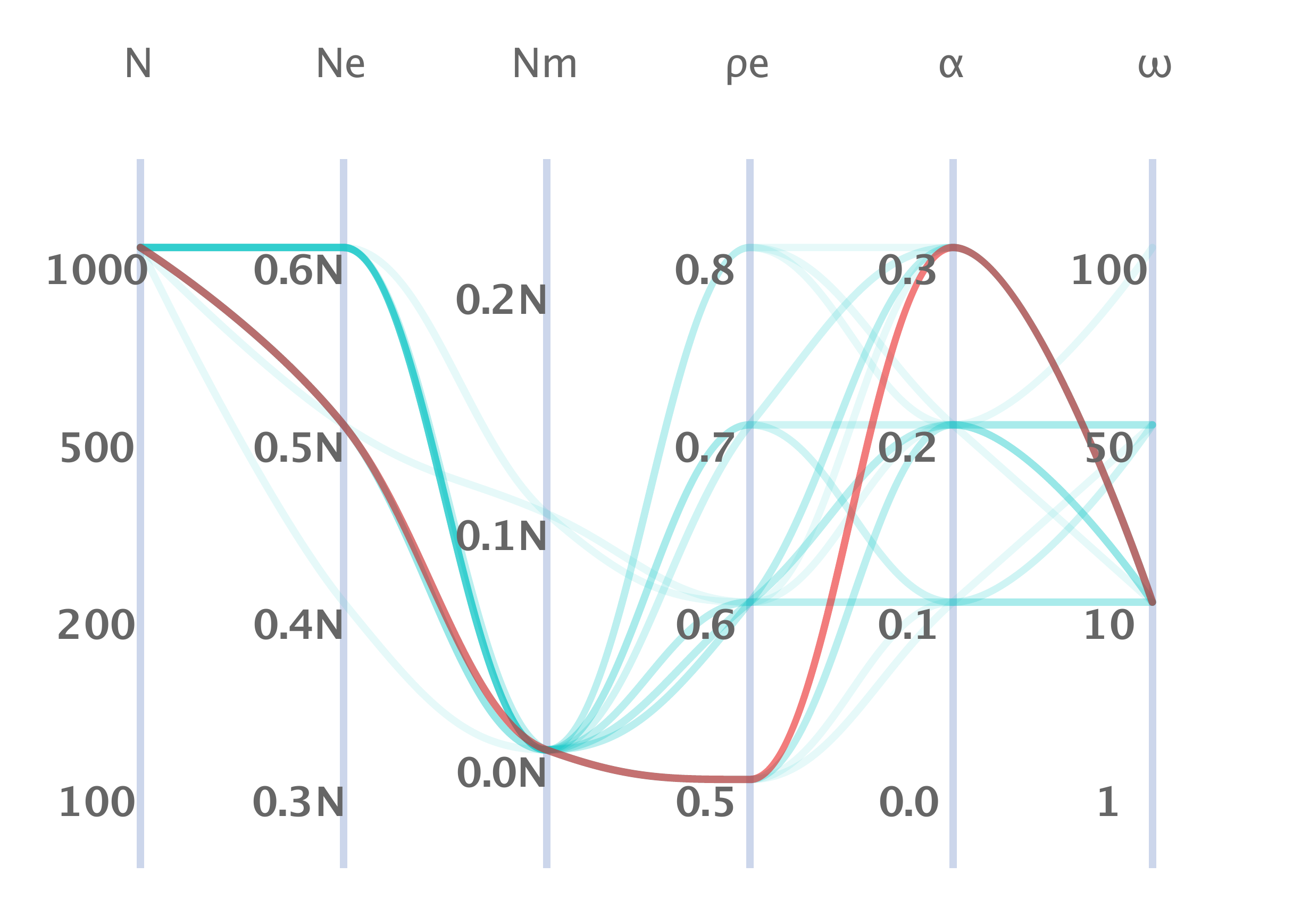}}
    
    \subfloat[7200 seconds]{\includegraphics[width=0.45\textwidth]{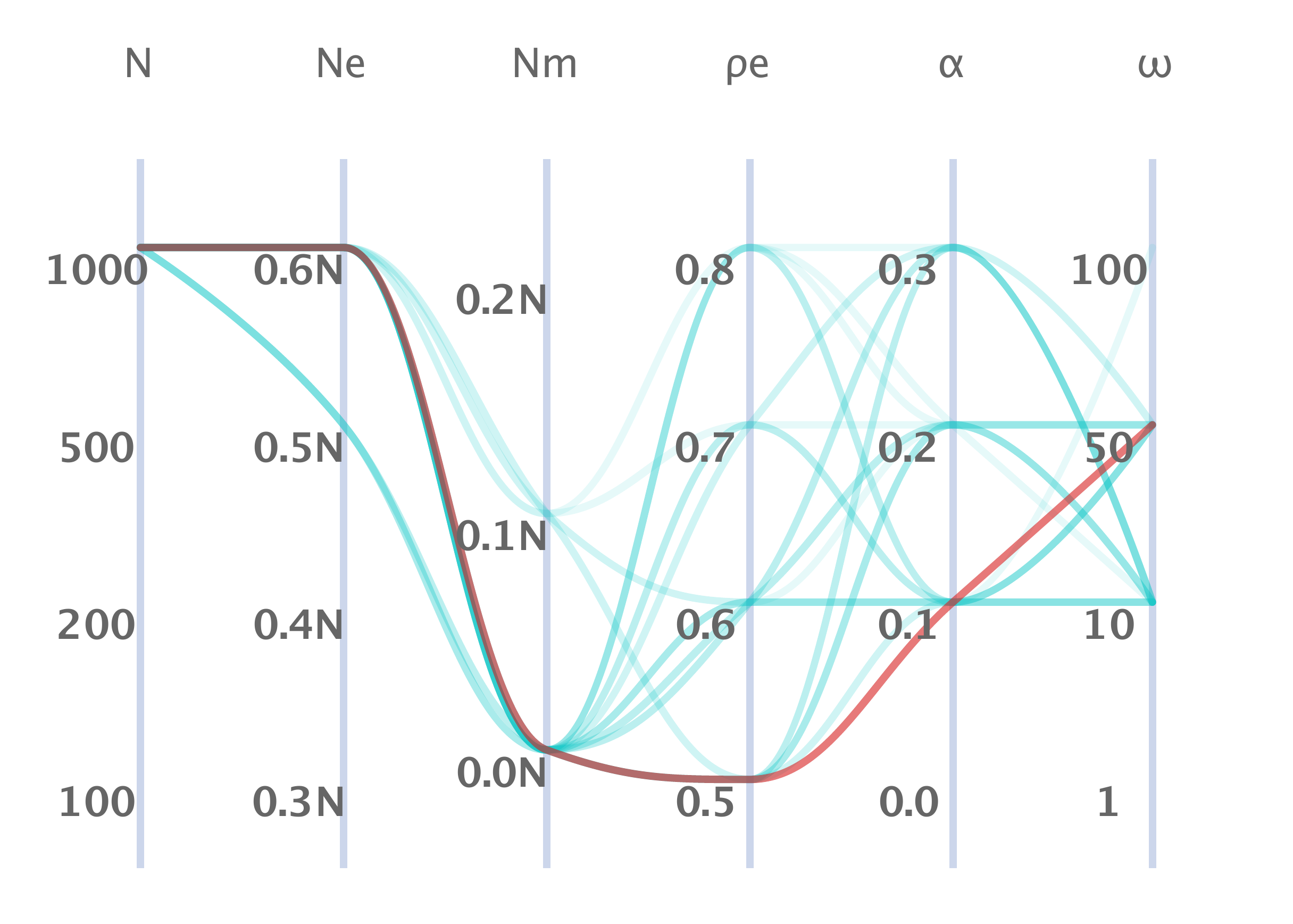}}
    \qquad \quad
    \subfloat[18000 seconds]{\includegraphics[width=0.45\textwidth]{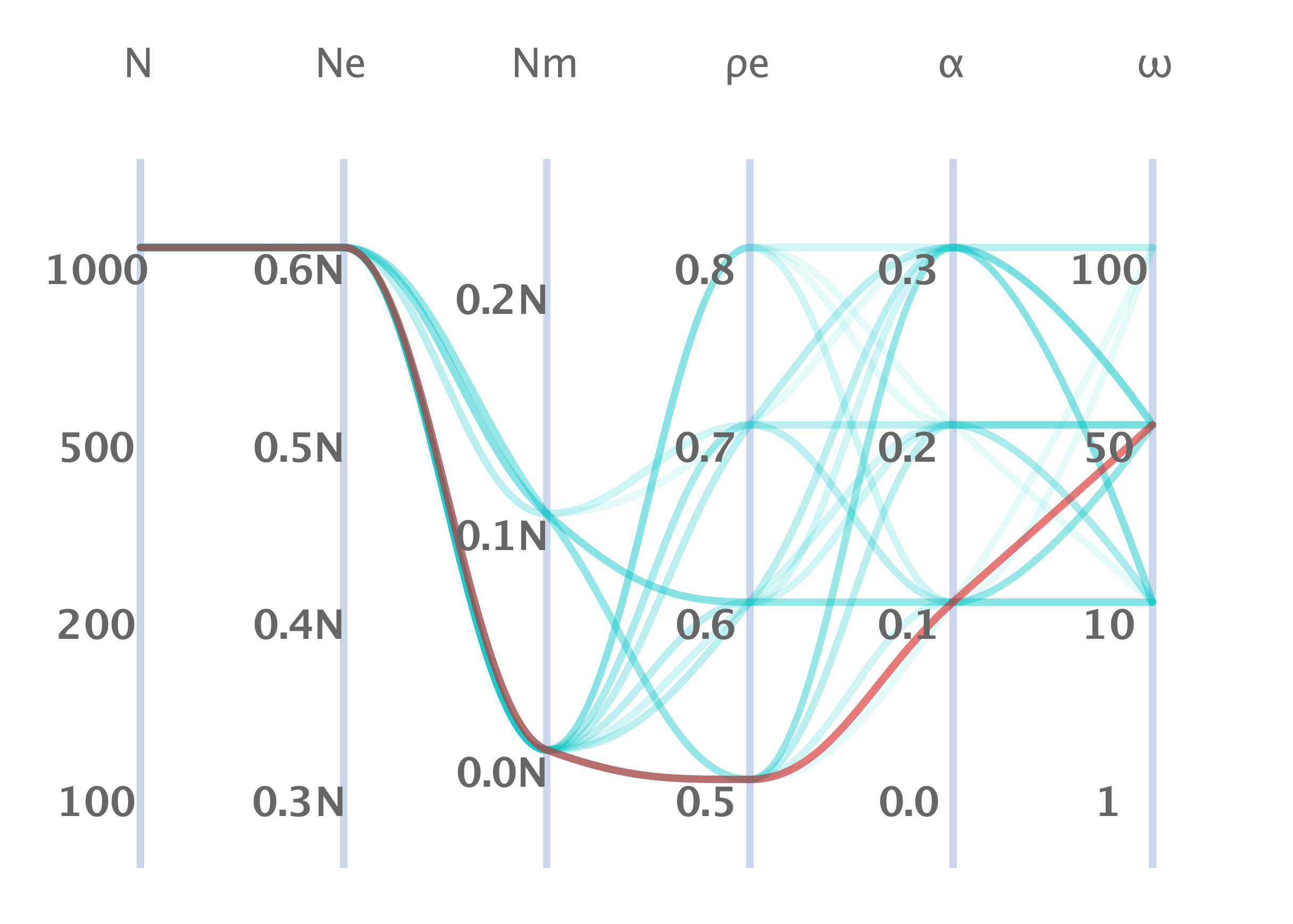}}
    \caption{Best parameter configurations over all instances with varying execution times.}
    \label{fig:parameter_configurations_parallel_axis}
\end{figure*}

In the following, we analyze the behavior of the parameters considering all instances together. In Figure~\ref{fig:parameter_configurations_parallel_axis}, we can visualize the best parameter configurations at six different execution times. In each plot, the best obtained parameter configuration regarding hypervolume is highlighted in red and parameter configurations up to 0.1\% worse than the best are highlighted in blue. Note that the intensity of the blue color indicates the importance of values of each parameter among the best parameter configurations once some parameter configurations share some parameter values. The following can be observed:

\begin{enumerate}[label=(\roman*), leftmargin=13mm]
\setlength\itemsep{2mm}

\item \textbf{More execution time, better results:}
The number of parameter configurations that are capable of generating large hypervolume values increases as the execution time of our algorithm increases. This means that in some runs, even though the parameters have not been set appropriately, the algorithm is still able to converge. 

\item \textbf{Importance of TSP and KP solvers:}
It influences the overall performance of the algorithm if TSP and KP solvers are used for initialization, which is determined by $\alpha$. The best results are obtained if at least $10\%$ percent of the initial solutions are biased towards those solutions found a TSP and KP solvers.

\item \textbf{Trends when execution time increases}: 
We can see a trend as the execution time increases. Our method performs better with a large population, a high survival rate, a small or no explicit diversification through mutant individuals, a small influence of single-parent inheritance, a minor influence of a good initial population, and significant influence of local search procedure. 
\end{enumerate}

\subsection{Competition Results}
\label{sec:results}

In order to analyze the efficiency of NDS-BRKGA compared to other methods, we present the results of the BI-TTP competitions held at \textit{EMO-2019} and \textit{GECCO-2019} conferences, where our method has been used and its solutions have been submitted. Both competitions did not have any regulations regarding running time, number of processors, or any other constraint and the ranking was solely based on the solution set submitted by each participant. Following the criteria of both competitions, we have compared the efficiency of the solutions of each submission for each test instance according to the hypervolume indicator. For each instance, the reference point used to calculate the hypervolume has been defined as the maximum time and the minimum profit obtained from the non-dominated solutions, which have been built from all submitted solutions.

In both competitions, the number of accepted solutions for each instance has been limited: For test instances based on \textit{a280} to $100$, \textit{fnl4461} to $50$, and for \textit{pla33810} to $20$. Because NDS-BRKGA returns a non-dominated set of solutions (named here as $A$), where its size can be larger than the maximum number of solutions $q$ accepted, we apply the dynamic programming algorithm developed by \cite{auger2009investigating}, which is able to find a subset $A^{*} \subseteq A$ with $\vert A^{*} \vert\, = q$, such that the weighted hypervolume indicator of $A^{*}$ is maximal. As stated by \cite{auger2009investigating}, this dynamic programming can be solved in time $\mathcal{O}(\vert A \vert ^{3})$.

For the \textit{EMO-2019} competition, we have used a preliminary version of the NDS-BRKGA described in Section \ref{sec:nsbrkga}. At that time, our method did not use the local search procedure described in Algorithm \ref{alg:self_improvement}. Moreover, the initial population of the algorithm used in that version has been created essentially at random. Only four individuals have been created from the TSP and KP solutions, which have been obtained by the same solvers previously described in this work. More precisely, those four individuals have been built from BI-TTP solutions $(\pi, \varnothing)$, $(\pi', \varnothing)$, $(\pi,z)$, and $(\pi', z)$, where $\pi$ is the tour found by LKH algorithm, $\pi'$ is the symmetric tour to $\pi$, and $z$ is the packing plan for the knapsack obtained by GH+DP algorithm.

In addition to our method, five other teams also submitted their solutions to the \textit{EMO-2019} competition. Among all submissions, NDS-BRKGA has had the best performance in seven of the nine test instances, resulting in the first-place award. All competition details, classification criteria, and results can be found at \url{https://www.egr.msu.edu/coinlab/blankjul/emo19-thief/}, where our submission is identified as \textit{``jomar''} (a reference to the two authors (\textbf{Jo}natas and \textbf{Mar}cone) who first worked on our algorithm). We herewith also present later a brief summary of these results.

After the \textit{EMO-2019} competition, we have realized that the inclusion of more individuals created from TSP and KP solvers helped the evolutionary process of our algorithm by combining more individuals with higher fitness from the first evolutionary cycles. Therefore, we initialize the population as shown in Algorithm \ref{alg:initial_population}. This new initial population initialization has been used in the \textit{GECCO-2019} competition, another BI-TTP competition that has considered the same criteria and classification rules of the \textit{EMO-2019} competition.

In the \textit{GECCO-2019} competition, $13$ teams have submitted their solutions. In this competition, NDS-BRKGA has won the second place in the final ranking. All detailed competition results can be found at \url{https://www.egr.msu.edu/coinlab/blankjul/gecco19-thief/}, where our submission is identified as \textit{``jomar''} again.

Table \ref{table:emo_gecco_results} shows a summary of the final results of both BI-TTP competitions. For each instance, we list the hypervolume achieved by the five best approaches that have been submitted to each competition. Our results from back then are highlighted in bold.

\begin{table}
\footnotesize
\centering
\caption{Results of the BI-TTP competitions. The results obtained by NDS-BRKGA were submitted by the team \textit{jomar}.}
\setlength{\tabcolsep}{0pt}
\begin{tabular*}{\hsize}{@{}@{\extracolsep{\fill}}ccrlcr@{}}
\toprule
\multirow{2}{*}{Instance} & \multicolumn{ 2}{c}{EMO-2019} &  & \multicolumn{ 2}{c}{GECCO-2019} \\ 
\cmidrule{2-3} \cmidrule{5-6}
\multicolumn{ 1}{c}{} & \multicolumn{1}{c}{Approach} & \multicolumn{1}{c}{HV} &  & \multicolumn{1}{c}{Approach} & \multicolumn{1}{c}{HV} \\ 
\midrule
\multirow{5}{*}{a280\_n279} & \textBF{jomar} & \textBF{0.893290} &  & HPI & 0.898433 \\ 
 & ALLAOUI & 0.835566 &  & \textBF{jomar} & \textBF{0.895567} \\ 
 & shisunzhang & 0.823563 &  & shisunzhang & 0.886576 \\ 
 & rrg & 0.754498 &  & NTGA & 0.883706 \\ 
 & CIRG\_UP\_KUNLE & 0.000000 &  & ALLAOUI & 0.873484 \\
\midrule
\multirow{5}{*}{a280\_n1395} & \textBF{jomar} & \textBF{0.816607} &  & HPI & 0.825913 \\ 
 & shisunzhang & 0.756445 &  & \textBF{jomar} & \textBF{0.821656} \\ 
 & rrg & 0.684549 &  & shisunzhang & 0.820893 \\ 
 & ALLAOUI & 0.581371 &  & NTGA & 0.811490 \\ 
 & CIRG\_UP\_KUNLE & 0.000000 &  & ALLAOUI & 0.808998 \\ 
\midrule
\multirow{5}{*}{a280\_n2790} & \textBF{jomar} & \textBF{0.872649} &  & \textBF{jomar} & \textBF{0.887945} \\ 
 & shisunzhang & 0.861102 &  & HPI & 0.887571 \\ 
 & rrg & 0.704428 &  & ALLAOUI & 0.885144 \\ 
 & ALLAOUI & 0.621785 &  & NTGA & 0.882562 \\ 
 & CIRG\_UP\_KUNLE & 0.000000 &  & shisunzhang & 0.874371 \\ 
\midrule
\multirow{5}{*}{fnl4461\_n4460} & \textBF{jomar} & \textBF{0.794519} &  & HPI & 0.933901 \\ 
 & shisunzhang & 0.719242 &  & \textBF{jomar} & \textBF{0.932661} \\ 
 & ALLAOUI & 0.553804 &  & NTGA & 0.914043 \\ 
 & CIRG\_UP\_KUNLE & 0.000000 &  & ALLAOUI & 0.889219 \\ 
 & OMEGA & 0.000000 &  & SSteam & 0.854150 \\ 
\midrule
\multirow{5}{*}{fnl4461\_n22300} & shisunzhang & 0.670849 &  & HPI & 0.818938 \\ 
 & \textBF{jomar} & \textBF{0.554188} &  & \textBF{jomar} & \textBF{0.814634} \\ 
 & ALLAOUI & 0.139420 &  & NTGA & 0.803470 \\ 
 & CIRG\_UP\_KUNLE & 0.000000 &  & SSteam & 0.781462 \\ 
 & OMEGA & 0.000000 &  & ALLAOUI & 0.760480 \\ 
\midrule
\multirow{5}{*}{fnl4461\_n44600} & shisunzhang & 0.540072 &  & HPI & 0.882894 \\ 
 & \textBF{jomar} & \textBF{0.534185} &  & \textBF{jomar} & \textBF{0.874688} \\ 
 & ALLAOUI & 0.009693 &  & SSteam & 0.856863 \\ 
 & CIRG\_UP\_KUNLE & 0.000000 &  & shisunzhang & 0.850339 \\ 
 & OMEGA & 0.000000 &  & NTGA & 0.824830 \\ 
\midrule
\multirow{5}{*}{pla33810\_n33809} & \textBF{jomar} & \textBF{0.718148} &  & HPI & 0.927214 \\ 
 & shisunzhang & 0.496913 &  & NTGA & 0.888680 \\ 
 & ALLAOUI & 0.090569 &  & ALLAOUI & 0.873717 \\ 
 & CIRG\_UP\_KUNLE & 0.000000 &  & \textBF{jomar} & \textBF{0.845149} \\ 
 & OMEGA & 0.000000 &  & SSteam & 0.832557 \\ 
\midrule
\multirow{5}{*}{pla33810\_n169045} & \textBF{jomar} & \textBF{0.697086} &  & HPI & 0.818259 \\ 
 & shisunzhang & 0.022390 &  & SSteam & 0.776638 \\ 
 & ALLAOUI & 0.007377 &  & NTGA & 0.773589 \\ 
 & CIRG\_UP\_KUNLE & 0.000000 &  & ALLAOUI & 0.769078 \\ 
 & OMEGA & 0.000000 &  & \textBF{jomar} & \textBF{0.738509} \\ 
\midrule
\multirow{5}{*}{pla33810\_n338090} & \textBF{jomar} & \textBF{0.696987} &  & HPI & 0.876129 \\ 
 & shisunzhang & 0.049182 &  & SSteam & 0.853805 \\ 
 & ALLAOUI & 0.001853 &  & \textBF{jomar} & \textBF{0.853683} \\ 
 & CIRG\_UP\_KUNLE & 0.000000 &  & ALLAOUI & 0.836965 \\ 
 & OMEGA & 0.000000 &  & NTGA & 0.781286 \\ 
\bottomrule
\end{tabular*}
\label{table:emo_gecco_results}
\end{table}

The results of the \textit{EMO-2019} competition show that the difference between the hypervolume achieved by NDS-BRKGA and by the others on smaller instances has been less significant than on larger instances. Also, it is worth mentioning that for the instances \textit{pla33810\_n169045} and \textit{pla33810\_n338090} the difference between the NDS-BRKGA and the second-best approach has been larger than 0.65 (65\% of the total hypervolume). 
Regarding the results of NDS-BRKGA and other submissions, we can clearly see the improvement achieved, especially for larger instances, by considering the current form of generating the initial population. The results of the instances \textit{fnl4461\_n22300} and \textit{fnl4461\_n44600} have been significantly improved compared to results obtained with the preliminary version of the algorithm submitted to the \textit{EMO-2019} competition.

\begin{figure}[!ht]
    \centering
    \includegraphics[width=\textwidth]{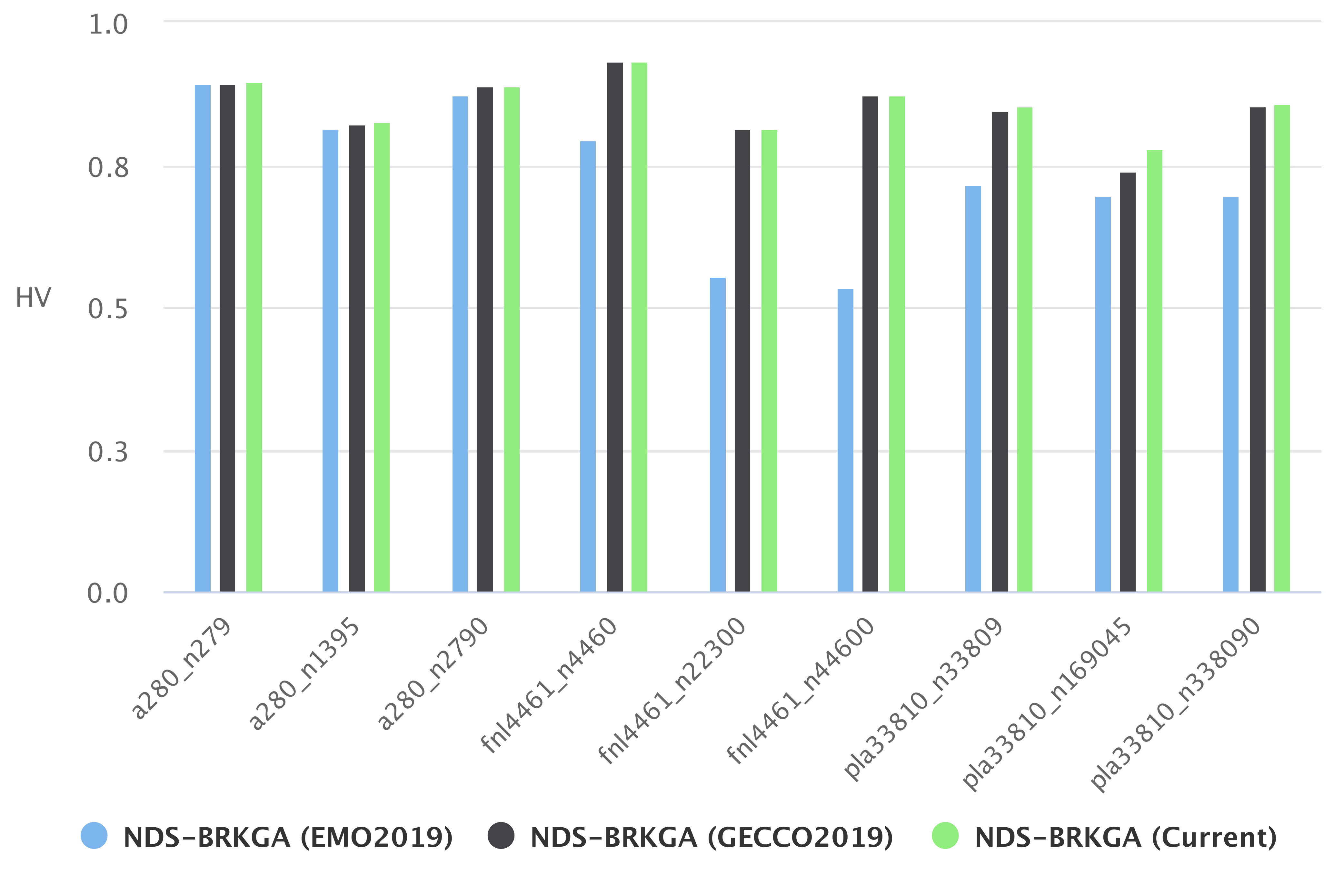}
    \caption{Hypervolumes obtained by the different NDS-BRKGA versions.}
    \label{fig:all_ndsbrkga_versions}
\end{figure}

The results of the competition \textit{GECCO-2019} show that NDS-BRKGA was able to win the test instance \textit{a280\_n2790} and has reached the second place five times. For test instances based on \textit{pla33810}, NDS-BRKGA was able to achieve one of the top five ranks ($11$ participants in total).
After the \textit{GECCO-2019} competition, we incorporated the exploitation phase as the most recent enhancement to our method, thus completing the NDS-BRKGA described in Section \ref{sec:nsbrkga}. In order to compare the performance of all versions, we plot the hypervolume reached by each version in each instance according to the criteria of the competitions (see Figure \ref{fig:all_ndsbrkga_versions}).
It can be observed that including more individuals from good solutions of the individual BI-TTP components brought a significant improvement. However, we did not observe major improvement after incorporating the exploitation phase after $5$ hours running time, except for the test instance \textit{pla33810\_n169045} in which the hypervolume increases around 4.2\%. 
Nevertheless, we have noticed a significant faster convergence with the incorporation. 

Because we have improved our method after the \textit{GECCO-2019} competition has passed, we have reevaluated the results based on the final version of NDS-BRKGA proposed in this paper. The results are shown in Table \ref{table:emo_gecco_ndsbrkga_results}.
In addition to the results, we present the hypervolume for each instance achieved by the four best approaches that have been submitted to both BI-TTP competitions. In the last column of the table, we list the difference between the hypervolume reached by each approach and that reached by the best one for each instance.

\begin{table}
\scriptsize
\centering
\caption{Best result of the BI-TTP competitions \textit{vs.} our final NDS-BRKGA.}
\setlength{\tabcolsep}{0pt}
\begin{tabular*}{\hsize}{@{}@{\extracolsep{\fill}}lcrr@{}}
\toprule
\textbf{Instance} & \textbf{Approach} & \multicolumn{1}{c}{\textbf{HV}} & \multicolumn{1}{c}{\textbf{Diff.}} \\ 
\midrule
\multirow{5}{*}{a280\_n279} & HPI & 0.898433 & 0.000000 \\ 
 & \textBF{NDS-BRKGA} & \textBF{0.895708} & \textBF{0.002725} \\ 
 & shisunzhang & 0.886576 & 0.011857 \\ 
 & NTGA & 0.883706 & 0.014727 \\ 
 & ALLAOUI & 0.873484 & 0.024949 \\
\midrule
\multirow{5}{*}{a280\_n1395}  & \textBF{NDS-BRKGA} & \textBF{0.826879} & \textBF{0.000000} \\ 
 & HPI & 0.825913 & 0.000966 \\ 
 & shisunzhang & 0.820893 & 0.005986 \\ 
 & NTGA & 0.811490 & 0.015389 \\ 
 & ALLAOUI & 0.808998 & 0.017881 \\
\midrule
\multirow{5}{*}{a280\_n2790}  & \textBF{NDS-BRKGA} & \textBF{0.887945} & \textBF{0.000000} \\ 
 & HPI & 0.887571 & 0.000374 \\ 
 & ALLAOUI & 0.885144 & 0.002801 \\ 
 & NTGA & 0.882562 & 0.005383 \\ 
 & shisunzhang & 0.874371 & 0.013574 \\
\midrule
\multirow{5}{*}{fnl4461\_n4460}  & \textBF{NDS-BRKGA} & \textBF{0.933942} & \textBF{0.000000} \\ 
 & HPI & 0.933901 & 0.000041 \\ 
 & NTGA & 0.914043 & 0.019899 \\ 
 & ALLAOUI & 0.889219 & 0.044723 \\ 
 & SSteam & 0.854150 & 0.079792 \\
\midrule
\multirow{5}{*}{fnl4461\_n22300} & HPI & 0.818938 & 0.000000 \\ 
 & \textBF{NDS-BRKGA} & \textBF{0.814492} & \textBF{0.004446} \\ 
 & NTGA & 0.803470 & 0.015468 \\ 
 & SSteam & 0.781462 & 0.037476 \\ 
 & ALLAOUI & 0.760480 & 0.058458 \\
\midrule
\multirow{5}{*}{fnl4461\_n44600}  & HPI & 0.882894 & 0.000000 \\ 
 & \textBF{NDS-BRKGA} & \textBF{0.874688} & \textBF{0.008206} \\ 
 & SSteam & 0.856863 & 0.026031 \\ 
 & shisunzhang & 0.850339 & 0.032555 \\ 
 & NTGA & 0.824830 & 0.058064 \\
\midrule
\multirow{5}{*}{pla33810\_n33809}  & HPI & 0.927214 & 0.000000 \\ 
 & NTGA & 0.888680 & 0.038534 \\ 
 & ALLAOUI & 0.873717 & 0.053497 \\ 
 & \textBF{NDS-BRKGA} & \textBF{0.852836} & \textBF{0.074378} \\ 
 & SSteam & 0.832557 & 0.094657 \\
\midrule
\multirow{5}{*}{pla33810\_n169045}   & HPI & 0.818259 & 0.000000 \\ 
 & \textBF{NDS-BRKGA} & \textBF{0.781009} & \textBF{0.037250} \\ 
 & SSteam & 0.776638 & 0.041621 \\ 
 & NTGA & 0.773589 & 0.044670 \\ 
 & ALLAOUI & 0.769078 & 0.049181 \\
\midrule
\multirow{5}{*}{pla33810\_n338090}  & HPI & 0.876129 & 0.000000 \\ 
 & \textBF{NDS-BRKGA} & \textBF{0.857105} & \textBF{0.019024} \\ 
 & SSteam & 0.853805 & 0.022324 \\ 
 & ALLAOUI & 0.836965 & 0.039164 \\ 
 & NTGA & 0.781286 & 0.094843 \\
 \bottomrule
\end{tabular*}
\label{table:emo_gecco_ndsbrkga_results}
\end{table}

One can notice that NDS-BRKGA has outperformed all other approaches in three instances (\textit{a280\_n1395}, \textit{a280\_n2790}, and  \textit{fnl4461\_n4460}). For the instances \textit{a280\_n279}, \textit{fnl4461\_n22300}, and \textit{fnl4461\_n44600}, NDS-BRKGA won the second place with a small difference to the first. For the three largest instances, NDS-BRKGA won the second place in two cases.

\subsection{Comparison with single-objective TTP solutions}\label{sec:comparisonSOTTP}

Lastly, we build the bridge to the single-objective TTP, which has been mostly investigated so far. 
Therefore, we compare our results with the single-objective TTP objective scores, which come from a comprehensive comparison of efficient algorithms already proposed in the literature. The computational budgets of the approaches which have obtained the best-known solutions might vary. 

Table \ref{table:bks_ttp_score_vs_ndsbrkga_ttp_score} compares for each instance the best-known score of the TTP with the best score found by our algorithm when it optimized the BI-TTP. Note that despite the strong connection of the BI-TTP to the single-objective TTP, maximizing the single-objective TTP objective score is not an explicit goal of the BI-TTP. Nevertheless, NDS-BRKGA has found better scores for the two smallest instances with $280$ cities with up to $1395$ items. In these cases, the best single-objective solutions are strictly dominated by our bi-objective solutions in both the bi-objective space and in the TTP score space.

\begin{table}
\footnotesize
\centering
\caption{Single-objective comparison of the TTP objectives scores.}
\begin{tabular*}{\hsize}{@{}@{\extracolsep{\fill}}lrr@{}}
\toprule
\multicolumn{1}{l}{\textbf{Instance}} & \multicolumn{1}{c}{\textbf{TTP score}$^{\star}$} & \multicolumn{1}{c}{\textbf{NDS-BRKGA}} \\ 
\midrule
a280\_n279 & 18526.000$^{a}$ & 18603.120 \\ 
a280\_n1395 & 112534.000$^{b}$ & 115445.521 \\ 
a280\_n2790 & 436932.000$^{b}$ & 429085.353 \\ 
fnl4461\_n4460 & 263040.254$^{c}$ & 257394.821 \\ 
fnl4461\_n22300 & 1705326.000$^{d}$ & 1567933.421 \\ 
fnl4461\_n44600 & 6744903.000$^{d}$ & 6272240.702 \\ 
pla33810\_n33809 & 1872169.000$^{c}$ & 1230174.003 \\ 
pla33810\_n169045 & 15707829.000$^{d}$ & 12935090.876 \\ 
pla33810\_n338090 & 58236645.120$^{e}$ & 55688288.508 \\ 
\bottomrule
\end{tabular*}
\begin{tablenotes}
\small
\item $^{\star}$ Best scores reported so far in the TTP articles, including in their supplementary files.
\item $^{a}$ HSEDA proposed by \cite{Martins2017ttpeda}; $^{b}$ MA2B proposed by \cite{el2016population}; $^{c}$ S5 proposed by \cite{faulkner2015approximate}; $^{d}$ CS2SA proposed by \cite{el2016population}; $^{e}$ C6 proposed by \cite{faulkner2015approximate}.
\end{tablenotes}
\label{table:bks_ttp_score_vs_ndsbrkga_ttp_score}
\end{table}

\begin{figure*}[!ht]
    \captionsetup[subfigure]{justification=centering, labelformat=empty}
    \centering
    \subfloat{\includegraphics[width=0.35\textwidth]{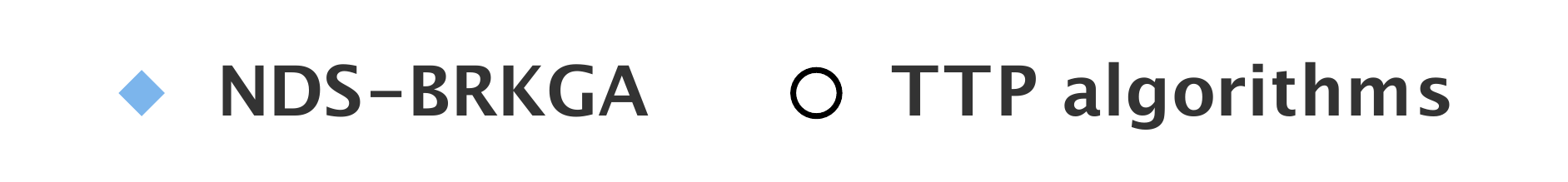}}
    
    \subfloat[a280\_n279 \text{[0.00\%]}]{\includegraphics[width=0.33\textwidth]{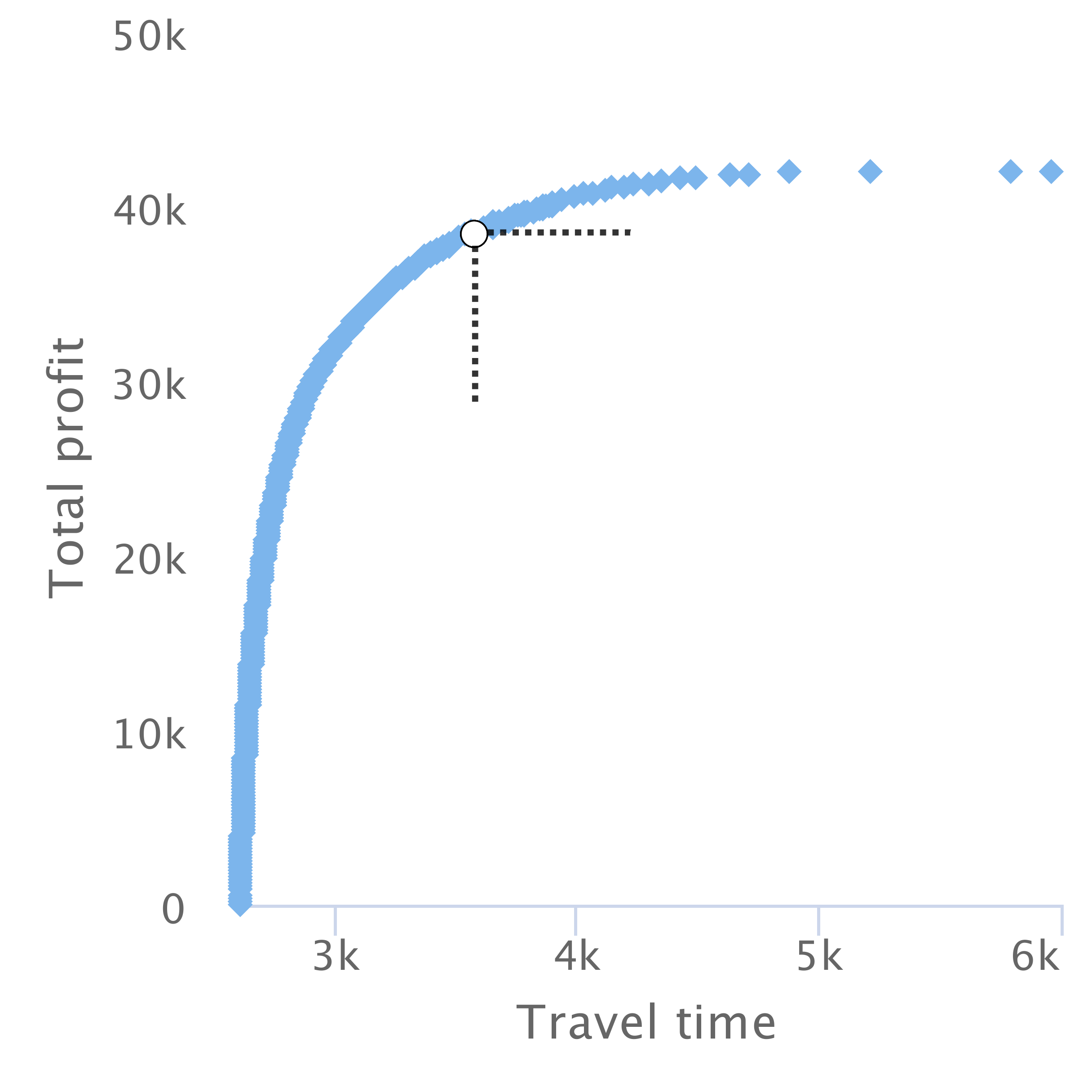}}
    \subfloat[a280\_n1395 \text{[0.00\%]}]{\includegraphics[width=0.33\textwidth]{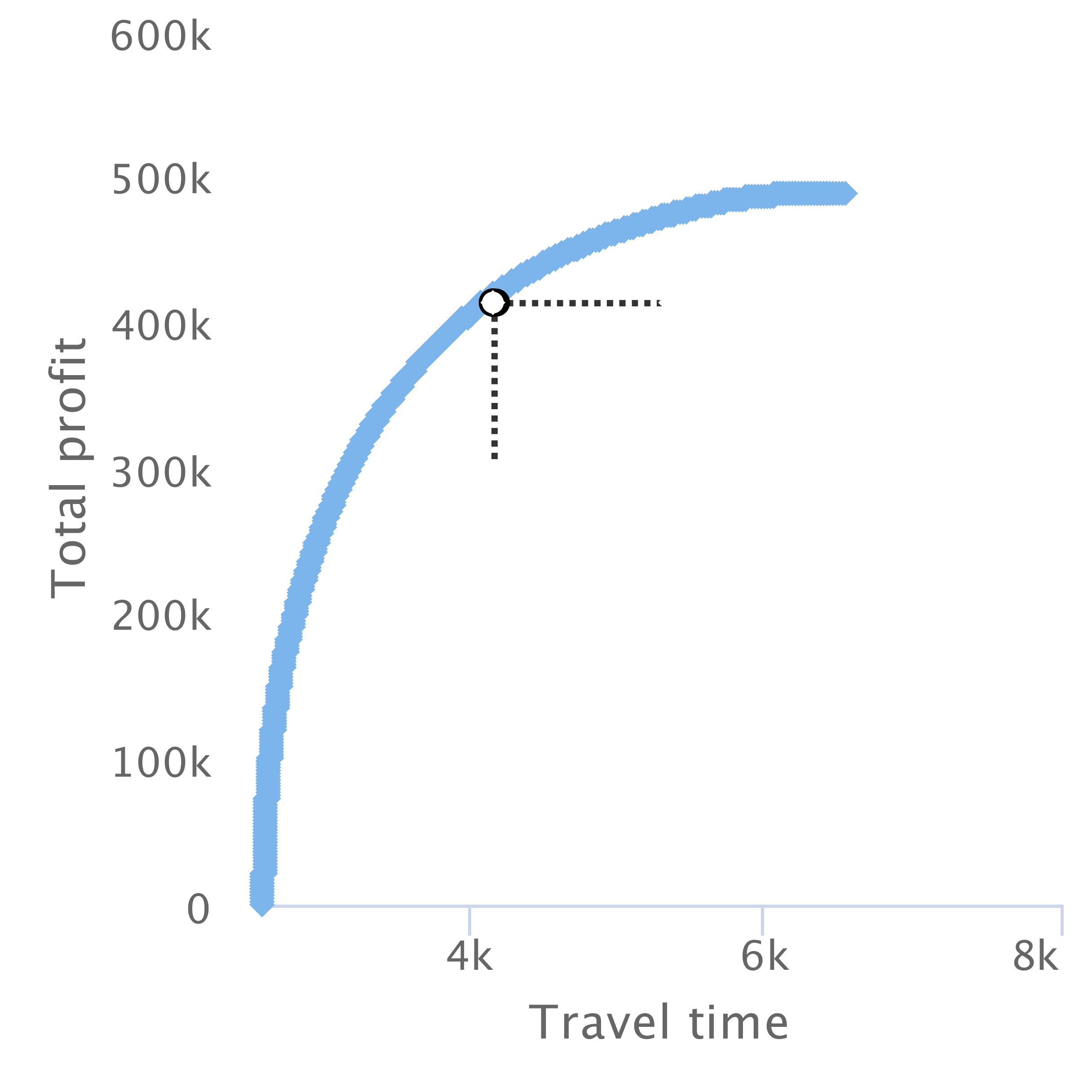}}
    \subfloat[a280\_n2790 \text{[0.31\%]}]{\includegraphics[width=0.33\textwidth]{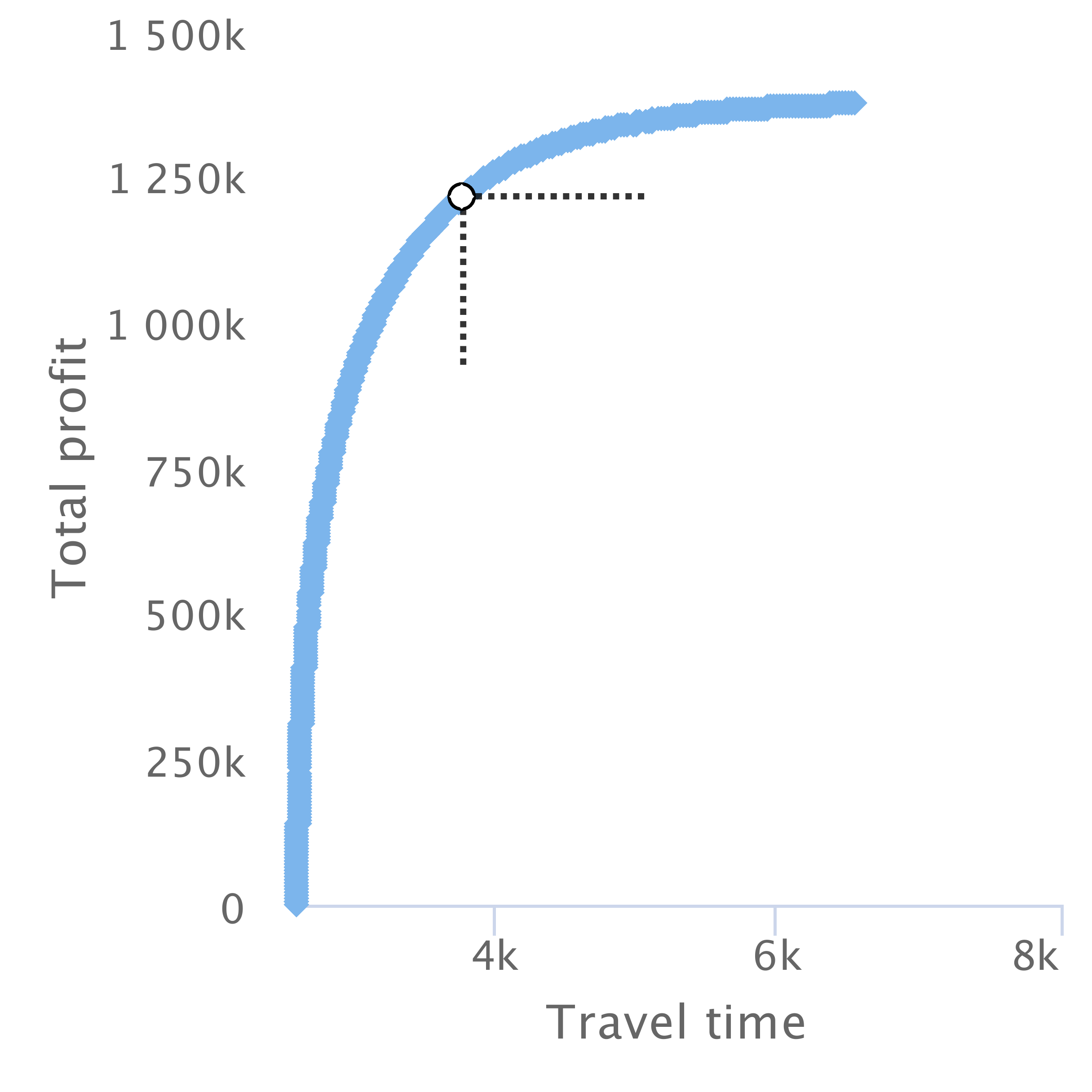}}
    
    \subfloat[fnl4461\_n4460 \text{[0.92\%]}]{\includegraphics[width=0.33\textwidth]{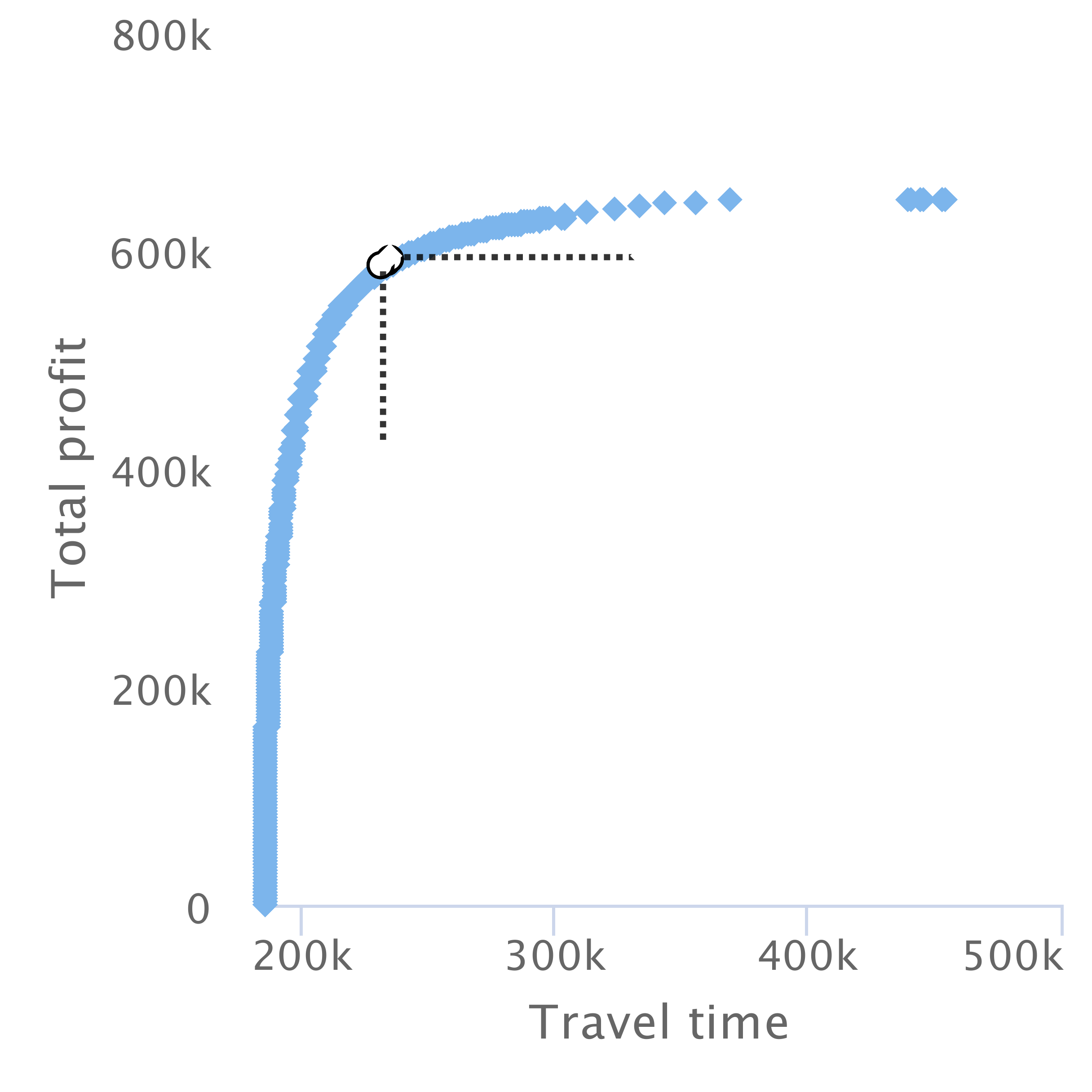}}
    \subfloat[fnl4461\_n22300 \text{[0.02\%]}]{\includegraphics[width=0.33\textwidth]{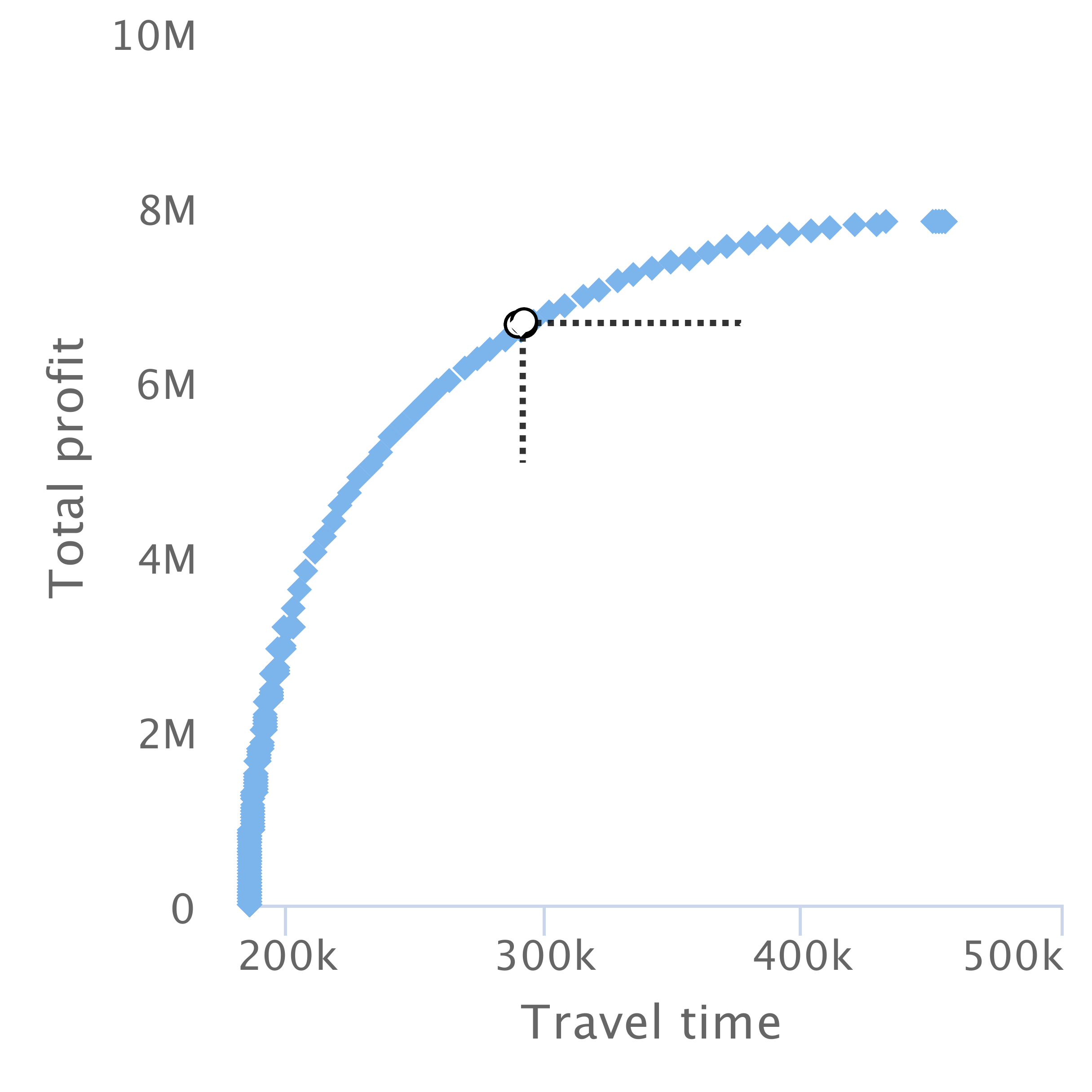}}
    \subfloat[fnl4461\_n44600 \text{[2.81\%]}]{\includegraphics[width=0.33\textwidth]{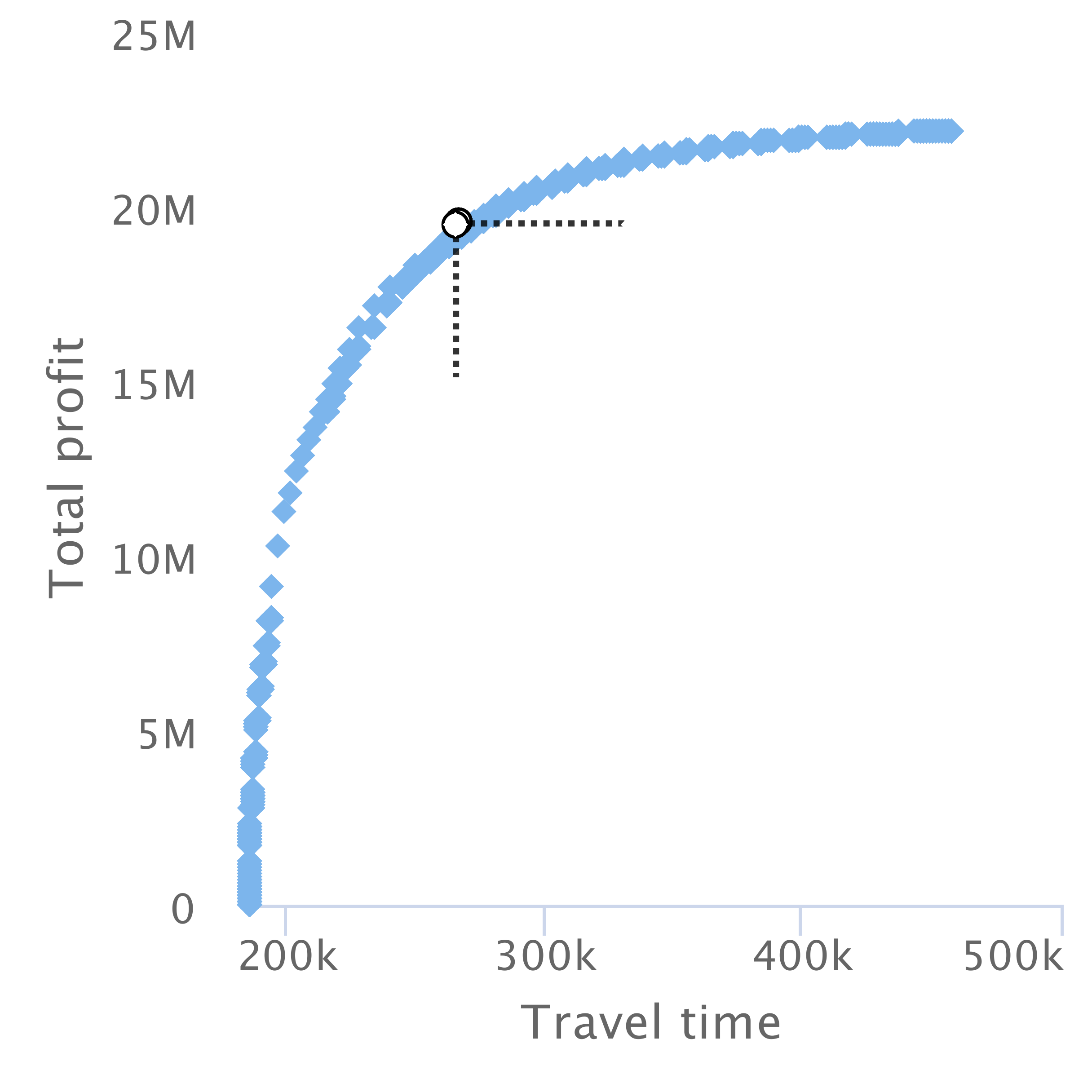}}
    
    \subfloat[pla33810\_n33809 \text{[18.93\%]}]{\includegraphics[width=0.33\textwidth]{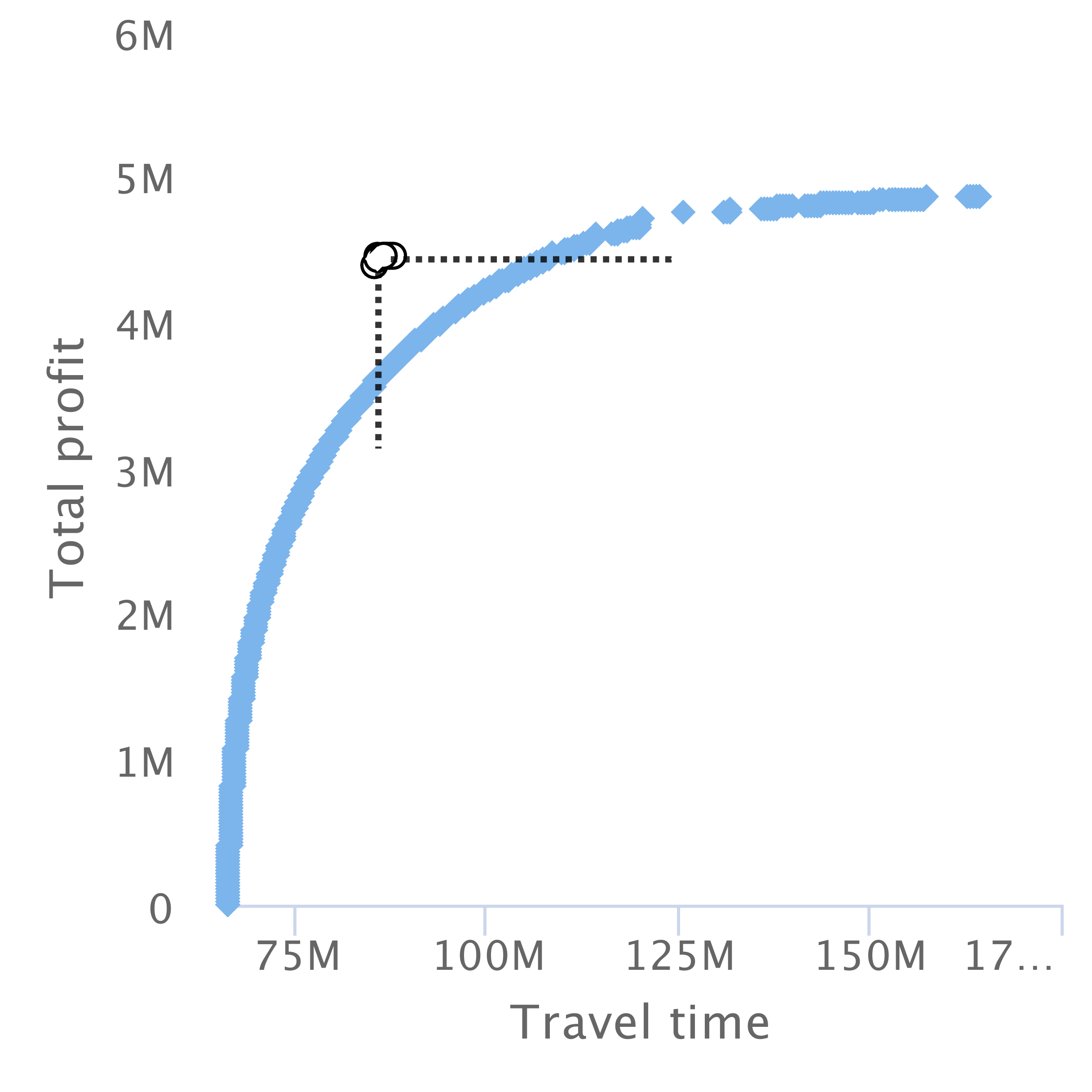}}
    \subfloat[pla33810\_n169045 \text{[4.11\%]}]{\includegraphics[width=0.33\textwidth]{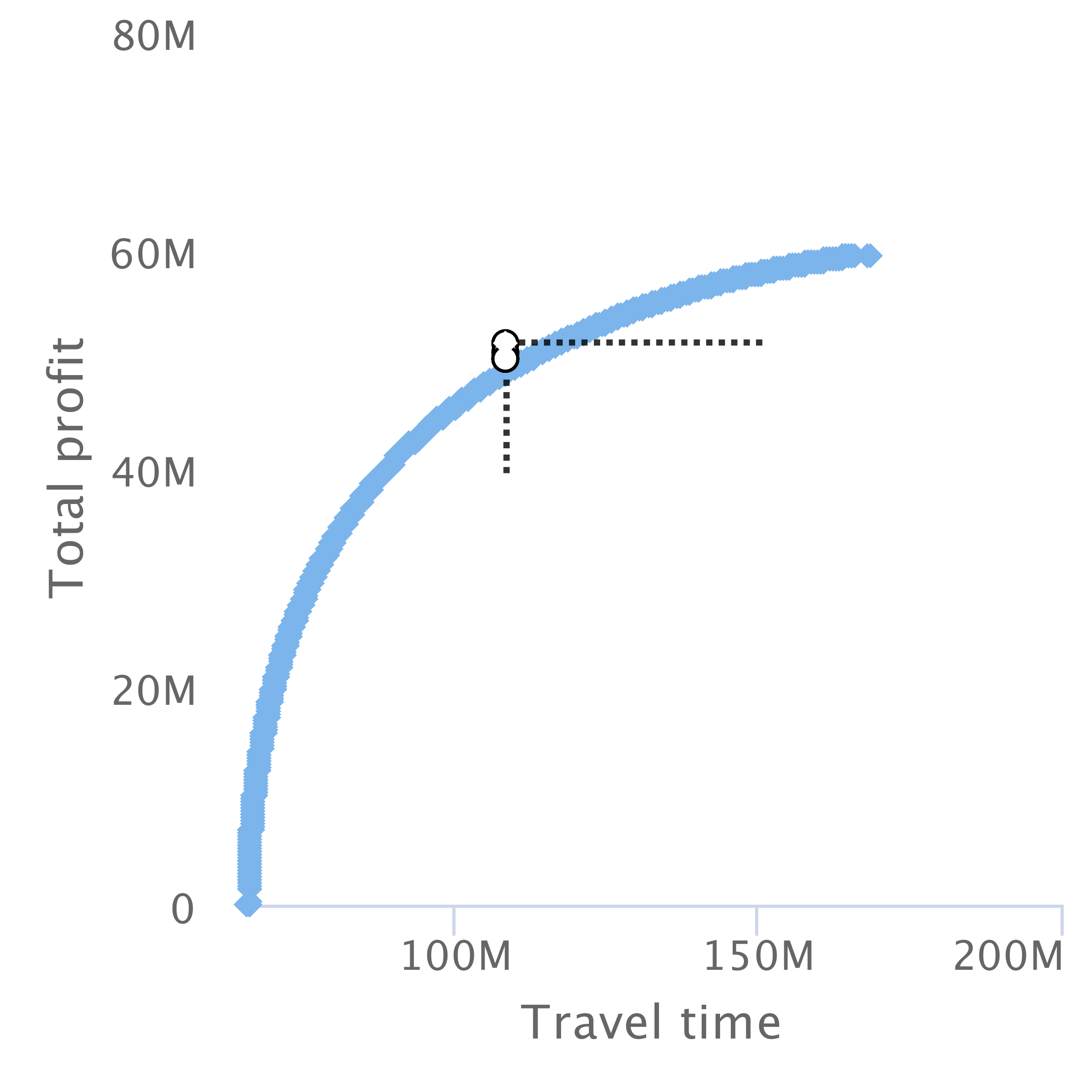}}
    \subfloat[pla33810\_n338090 \text{[1.36\%]}]{\includegraphics[width=0.33\textwidth]{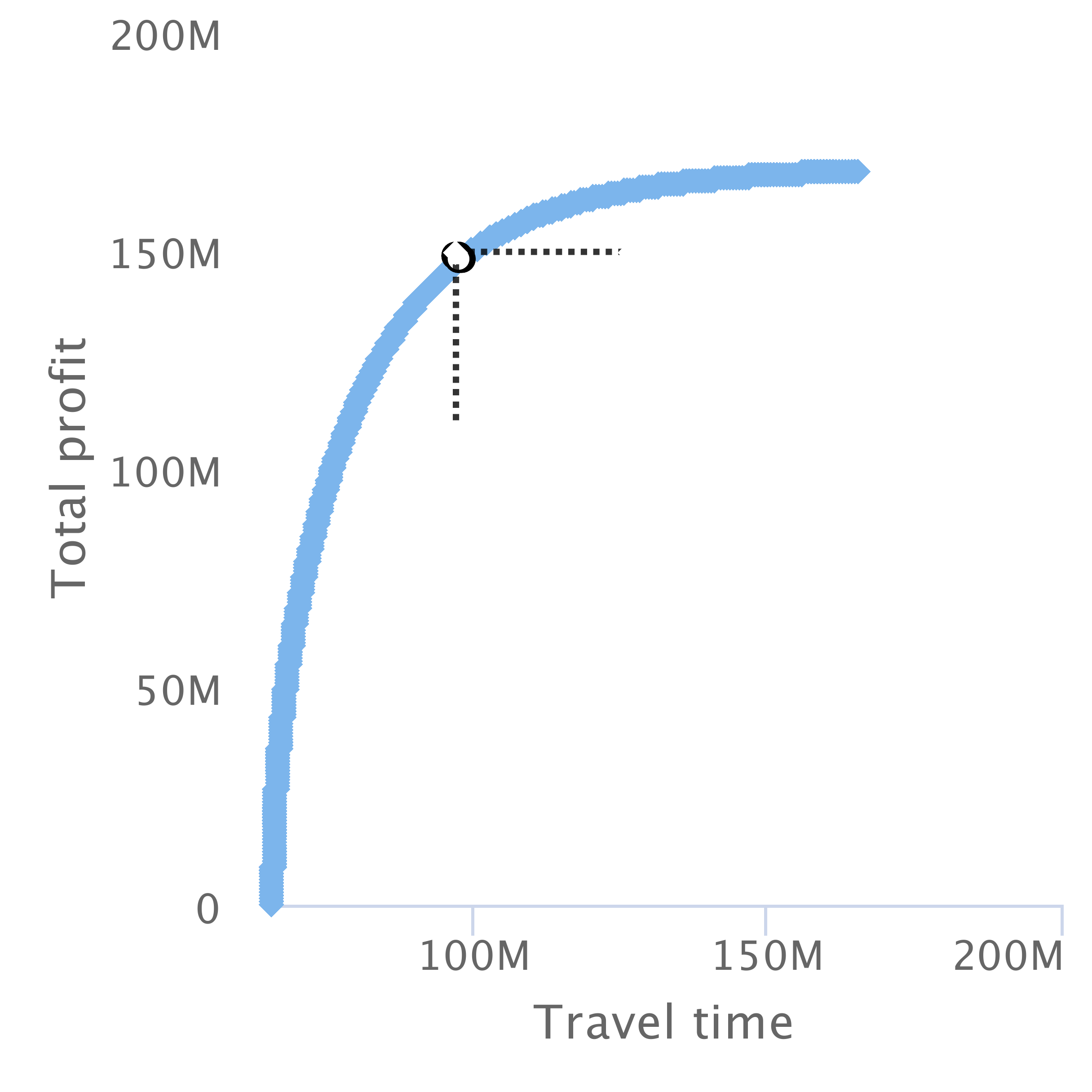}}
    \caption{NDS-BRKGA solutions for the present BI-TTP, and solutions generated by single-objective TTP algorithms. The numbers in brackets show the percentage of the bi-objective solutions that are dominated by the single-objective solutions.}
    \label{fig:bks_ttp_vs_ndsbrkga}
\end{figure*}

Figure~\ref{fig:bks_ttp_vs_ndsbrkga} shows the 100\% attainment surface for each instance. For each problem all non-dominated solutions found by our algorithm and the single-objective TTP solution obtained by the best performing algorithm (out of 10 runs) are plotted. Moreover, the dotted lines represent the dominated region of the solution obtained by a single-objective optimizer.
The figure clearly shows that for almost all instances, none or only a few solutions are dominated (see the values in brackets in the figure). Not only this, but also the fact that not a single but multiple non-dominated solutions have been obtained shows the efficiency of our proposed approach. Only for the problem instance \textit{pla33810\_n33809} the single-objective optimizer has been able to find significantly better results, where the single-objective solutions dominate 18.93\% of the bi-objective solutions.

\section{Concluding Remarks}
\label{sec:conclusion}

In this paper, we have investigated the bi-objective traveling thief problem where the Traveling Salesman and Knapsack Problem interact with each other. We have proposed an evolutionary optimization algorithm that uses the principles of customization to solve this challenging combinatorial optimization problem effectively. Each customization addresses one specific problem characteristic that needs either to be considered during the optimization or can be used to further improve the convergence of the algorithm.

Our proposed method has incorporated problem knowledge by creating a biased initial population that contains individuals generated by existing efficient solvers for each subproblem independently. Moreover, the constraint has been handled through a customized repair operator during the evolution, and the heterogeneous variables have been unified through a genotype to phenotype mapping. To address the existence of two objectives, we have used non-dominated sorting and crowding distance during the environment survival and to further improve the convergence, a local search has been applied selectively during evolution.

Since these customizations have come with a few parameters, we have conducted an extensive experiment to show the effect of each parameter on the overall performance of the algorithm. Our results have indicated that the best-performing configurations are those with larger population size, a higher survival rate for the best individuals, and a high contribution of the TSP and KP solvers for creating a part of the initial population. The contribution of mutant individuals has been found to be insignificant. 

As a future study, new ways of initializing the population is worth investigating. So far, we have used solutions obtained by subproblem solvers, but did not consider seeding it with good already-known TTP solution. Moreover, we are planning to efficiently incorporate the algorithmic insights gained from the single-objective approaches into the multi-objective setting; a naive bi-level approach is known to be computationally impractical.
Lastly, it is worth investigating how the proposed concepts can be used for other optimization problems where two problems interact with each other. This requires extending the proposed concepts to interwoven optimization problems in general and evaluating the method's generalizability.

\begin{acknowledgements}
The authors thank Coordena\c{c}\~{a}o de A\-per\-fei\-\c{c}o\-a\-men\-to de Pessoal de N\'{i}vel Superior (CAPES) - Finance code 001, Funda\c{c}\~{a}o de Amparo \`{a} Pesquisa do Estado de Minas Gerais (FAPEMIG), Conselho Nacional de Desenvolvimento Cient\'{i}fico e Tecnol\'{o}gico (CNPq), Universidade Federal de Ouro Preto (UFOP), Universidade Federal de Viçosa (UFV) for supporting this research. The authors would also like to thank the HPI Future SOC Lab and its team for enable this research by providing access to their compute infrastructure, \url{https://www.hpi.de/future-soc-lab}.
\end{acknowledgements}

%
%

\FloatBarrier
%
%
\bibliographystyle{spbasic}      

\bibliography{references}   

\end{document}